\documentclass[sigconf]{acmart}

\usepackage{xspace}
\usepackage{multirow}
\usepackage{placeins}
\usepackage{pifont}
\usepackage{balance}
\AtBeginDocument{%
  }

\copyrightyear{2026}
\acmYear{2026}
\setcopyright{cc}
\setcctype{by}
\acmConference[KDD '26]{Proceedings of the 32nd ACM SIGKDD Conference on Knowledge Discovery and Data Mining V.2}{August 09--13, 2026}{Jeju Island, Republic of Korea}
\acmBooktitle{Proceedings of the 32nd ACM SIGKDD Conference on Knowledge Discovery and Data Mining V.2 (KDD '26), August 09--13, 2026, Jeju Island, Republic of Korea}
\acmDOI{10.1145/3770855.3817523}
\acmISBN{979-8-4007-2259-2/2026/08}

\begin{document}

\title{Benchmark Datasets for Lead-Lag Forecasting on Social Platforms}

\author{Kimia Kazemian}
\authornote{Both authors contributed equally to this research.}
\email{kk983@cornell.edu}
\author{Zhenzhen Liu}
\authornotemark[1]
\email{zl535@cornell.edu}
\author{Yangfanyu Yang}
\email{yy2288@cornell.edu}
\affiliation{%
  \institution{Cornell University}
  \city{Ithaca}
  \state{NY}
  \country{USA}
}

\author{Katie Luo}
\email{katieluo@stanford.edu}
\affiliation{%
  \institution{Stanford University}
  \city{Palo Alto}
  \state{CA}
  \country{USA}
}

\author{Shuhan Gu}
\email{han0705@bu.edu}
\author{Audrey Du}
\email{md964@cornell.edu}
\affiliation{%
  \institution{Cornell University}
  \city{Ithaca}
  \state{NY}
  \country{USA}
}

\author{Xinyu Yang}
\email{xy468@cornell.edu}
\author{Jack Jansons}
\email{jcj59@cornell.edu}
\affiliation{%
  \institution{Cornell University}
  \city{Ithaca}
  \state{NY}
  \country{USA}
}

\author{Kilian Q. Weinberger}
\email{kqw4@cornell.edu}
\author{John Thickstun}
\email{jthickstun@cornell.edu}
\affiliation{%
  \institution{Cornell University}
  \city{Ithaca}
  \state{NY}
  \country{USA}
}

\author{Yian Yin}
\email{yy994@cornell.edu}
\author{Sarah Dean}
\email{sdean@cornell.edu}
\affiliation{%
  \institution{Cornell University}
  \city{Ithaca}
  \state{NY}
  \country{USA}
}

\renewcommand{\shortauthors}{Kazemian et al.}

\newcommand{\arxiv}{arXiv\xspace}
\newcommand{\github}{GitHub\xspace}

\begin{abstract}
Social and collaborative platforms emit multivariate time-series traces in which early interactions---such as views, likes, or downloads---are followed, sometimes months or years later, by higher impact like citations, sales, or reviews. We formalize this setting as \textbf{Lead-Lag Forecasting (LLF): given an early usage channel (the lead), predict a correlated but temporally shifted outcome channel (the lag).} Despite the ubiquity of such patterns, LLF has not been treated as a unified forecasting problem within the time-series community, largely due to the absence of standardized datasets. To anchor research in LLF, here we present two high-volume benchmark datasets: arXiv (accesses → citations of 2.3M papers) and GitHub (pushes/stars → forks of 3M repositories).
Our datasets provide ideal testbeds for lead–lag forecasting, by capturing long-horizon dynamics across years, spanning the full spectrum of outcomes, and avoiding survivorship bias in sampling.
We documented all technical details of data curation and cleaning, verified the presence of lead-lag dynamics through statistical and classification tests, and benchmarked parametric and non-parametric baselines for regression. Our study establishes LLF as a novel forecasting paradigm and lays an empirical foundation for its systematic exploration in social and usage data.
\end{abstract}




\newcommand\blfootnote[1]{%
  \begingroup
  \renewcommand\thefootnote{}\footnote{#1}%
  \addtocounter{footnote}{-1}%
  \endgroup
}
\maketitle

\section{Introduction}

The success of human activities is often measured by their collective impact, ranging from music streams and movie box office revenues to product sales and social media popularity. These impact metrics typically follow heavy-tailed distributions \citep{clauset2009power} and slow decay patterns across timescales \citep{candia2019universal}, making early identification of future hits fundamentally challenging \citep{cheng2014can,martin2016exploring}. At the same time, digital platforms increasingly log online user interactions---searches, views, downloads, likes, and shares---that often precede these long-term dynamics. These temporal lead-lag dynamics are remarkably ubiquitous, spanning domains as diverse as science \citep{haque2009positional,brody2006earlier}, economics \citep{wu2015future}, arts \citep{goel2010predicting}, culture \citep{gruhl2005predictive}, and social movements \citep{johnson2016new}.
A systematic understanding of such lead-lag dynamics is not only crucial for anticipating and optimizing impact in digital ecosystems, but also essential for designing effective strategies that identify and promote emerging innovations and products.

{
\newcommand{\cmark}{\ding{51}}
\newcommand{\xmark}{\ding{55}}
\begin{table*}[htbp]
    \centering
    \scriptsize
    \caption{\textbf{Comparison of time series datasets.}}
    \vspace{-2mm}
      \resizebox{0.9\textwidth}{!}
      { 
    \begin{tabular}{lp{25mm}cccccc}\toprule
        \multirow{2}{*}{\textbf{Name}} & 
        \multirow{2}{*}{\textbf{Domain}} & 
        \textbf{Labeled} & 
        \textbf{Multi.} & 
        \multirow{2}{*}{\textbf{\# Series}} & 
        \multicolumn{2}{c}{\textbf{Time Series Task}} &
        \textbf{Sugg. Pred.} \\
        \cmidrule(lr){6-7}
        & & \textbf{Lag Event?} & \textbf{Variate?} & & \textbf{Forecast} & \textbf{TTP} & \textbf{Horizon} \\\midrule
        Electricity & Energy & \xmark & \xmark & 370 & \cmark &  & 1 day \\
        Solar & Energy & \xmark & \xmark & 137 & \cmark & & 4 hrs \\
        ETT & Energy & \xmark & \cmark & 69 & \cmark & & 30 days \\
        Weather & Weather & \xmark & \cmark & 20 & \cmark & & 30 day \\
        NYC Taxi & Transport. & \xmark & \cmark & 12M & \cmark & \cmark & \textit{N/A}\\
        UberTLC & Transport. & \xmark & \cmark & 200K & \cmark & \cmark & \textit{N/A} \\
        Traffic & Transport. & \xmark & \xmark & 862 & & \cmark & 1 day \\
        KDD Cup & Security & \xmark & \cmark & 5M & \cmark & \cmark & \textit{N/A} \\
        Exchange & Economics & \xmark & \xmark & 8 & \cmark & & 24 days \\
        Wiki Traffic & Network & \xmark & \xmark & 145K & \cmark & \cmark & \textit{N/A} \\
        Retweet & Social & \xmark & \cmark & 166K & \cmark & \cmark & 6 hr\\
        M1--5 & M Competitions & \xmark & \xmark & 1--100K & \cmark & &  $<$ 1 yr \\
        \midrule
        Ours (arXiv) & Science & \cmark & \cmark & 2.3M & \cmark & \cmark & 5 yrs\\
        Ours (GitHub) & Technology & \cmark & \cmark & 3M & \cmark & \cmark & 5 yrs\\\bottomrule
        
    \end{tabular}
    }
    \label{tab:datasets}
\end{table*}
}
In this paper, we introduce \textbf{Lead-Lag Forecasting (LLF)} defined as follows: given a “lead” usage channel, predict a correlated but temporally shifted “lag” outcome channel. LLF builds upon early-signal prediction but sharpens the setting to one where the goal is not generic long-horizon forecasting, but recovering systematic temporal shifts between coupled signals—requiring the predictor to transfer information across channels (e.g., accesses → citations) while generalizing to new series.

LLF problems exhibit three key “dual process” properties:
(i) \textit{observable early-phase interactions}---e.g., views, likes, GitHub repository pushes,
(ii) \textit{meaningful outcome signal}---e.g., paper citations, purchases, GitHub repository forks, and 
(iii) \textit{predictive dependency}---the early dynamics often \textbf{cause or forecast} later outcomes through complex, non-deterministic mechanisms.
Despite the ubiquity of such dual process dynamics across digital ecosystems,
our quantitative understanding of their predictability remains limited, largely due to the lack of standardized datasets for systematic LLF research.
Though frequently recorded in internal web logs, early ``lead'' signals are not often made publicly available.
Popular forecasting benchmarks---traffic~\citep{caltrans2025pems}, taxi demand~\citep{nyctlc2025tripdata}, ETTh~\citep{zhou2021informer}, M4~\citep{iif2025timeseries}, and related corpora---have spurred substantial progress on short-horizon and seasonal forecasting within a single measurement channel. While these benchmarks serve their intended applications well, they focus primarily on same-channel prediction rather than cross-channel dynamics with series generalization.

To address this gap, here we build and release\footnote{Access to data, codes, and supplementary information can be found on our project website: \url{https://lead-lag-forecasting.github.io}.} two high-volume datasets in science (accesses and citations of about 2.3M \textbf{arXiv} papers) and technology (pushes, stars, and forks of 3M \textbf{GitHub} repositories).
These two examples represent ideal testbeds for LLF research: (1)  The broad impact of papers and software often extends beyond their original communities, making them significant, real-world settings rather than artificial test cases \citep{yin2022public}.
(2) Both systems
have observable effects that unfold over years, presenting complex real-world challenges that traditional methods struggle to model
\citep{ke2015defining, jiang2021hints}.
(3) As our records include both the high volume of low-impact items and the small number of high-impact breakthroughs, the datasets provide unbiased coverage across the entire impact spectrum without survivorship bias.

To summarize, our contributions are the following: (1) we provide two datasets, one of which has never before been publicly available, and document technical details of data curation and cleaning; (2) we analyze the lead-lag dynamics in both datasets and demonstrate predictive signals in the paired channels; (3) 
we benchmark deep learning, parametric, and non-parametric prediction baselines. Together, this study establishes LLF as a novel forecasting paradigm and lays an empirical foundation for its systematic exploration in social data.

\begin{figure*}[t]
    \centering
    \includegraphics[width=0.28\linewidth]{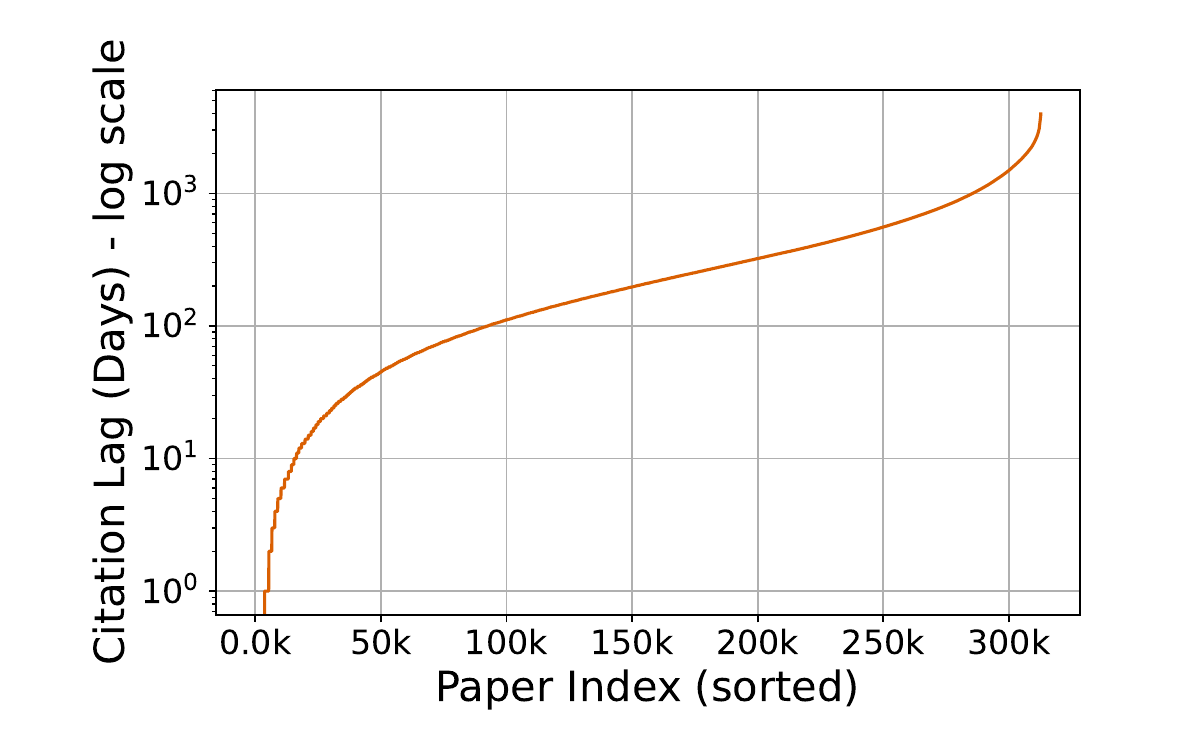}%
    \hspace{0.007\linewidth}%
    \includegraphics[width=0.23\linewidth]{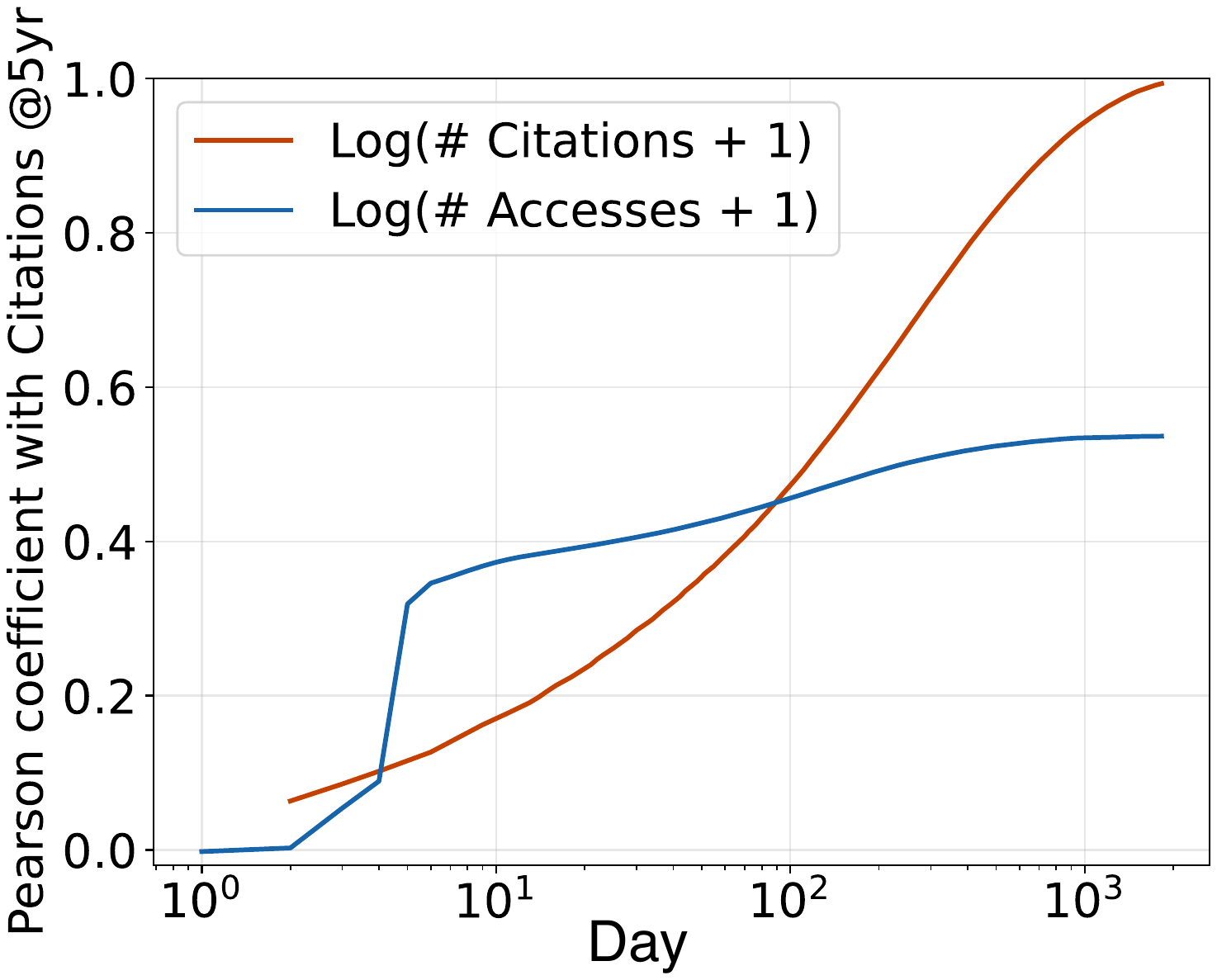}%
    \hspace{0.007\linewidth}%
    \includegraphics[width=0.22\linewidth]{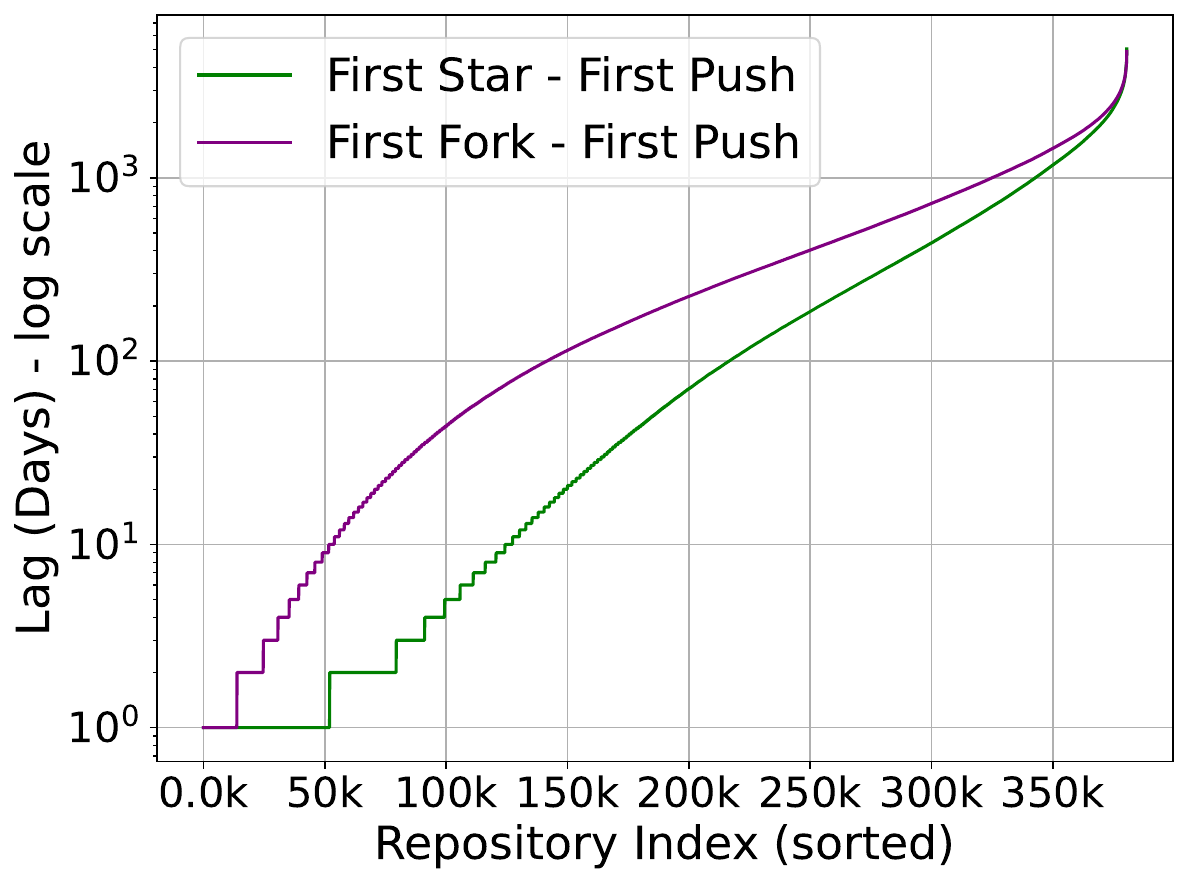}%
    \hspace{0.007\linewidth}%
    \includegraphics[width=0.24\linewidth]{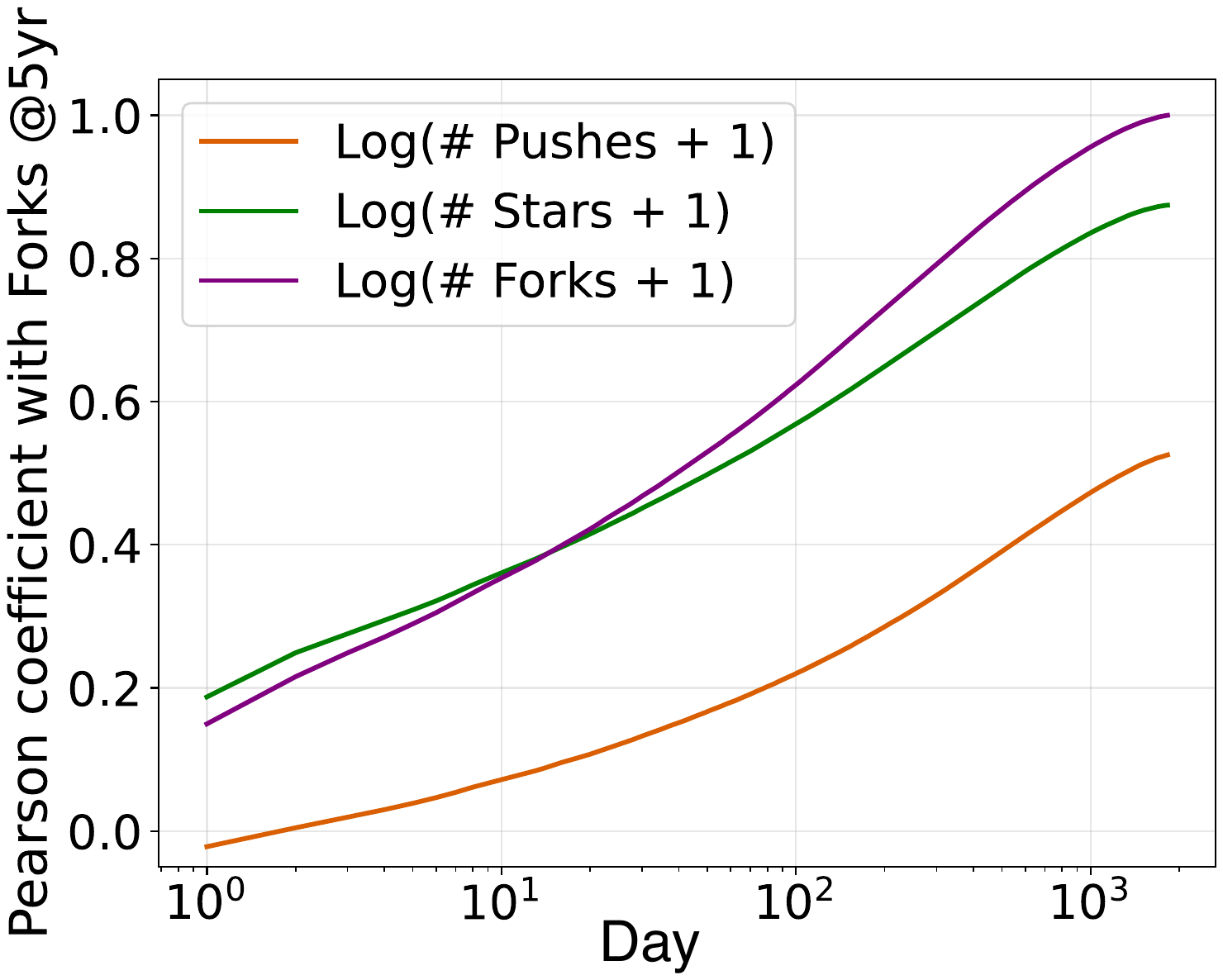}
    
    \vspace{3pt}
    
    \makebox[0.51\linewidth]{\textbf{arXiv}}\hfill\makebox[0.46\linewidth]{\textbf{GitHub}}
    \caption{\textbf{Left to right:} 1. Citation delay in days after first access. 2. Pearson correlation of early accesses (blue) and citations (orange) with 5-year citation count. Access data shows strong early correlation. After 3 months (90 days), citations gradually become more predictive. 3. Lag between first activity and first star and first fork event. 4. Pearson correlation of early pushes (orange), stars (green) and forks (blue) with 5-year fork count. Stars have the strongest correlation in the first 2 weeks. Afterwards, forks gradually become more predictive.}
    \label{fig:combined_analysis}
\end{figure*}
\section{Related Work}
\label{sec:related}
\textbf{Time Series Prediction Tasks.}
Time series prediction is a longstanding and active research area in machine learning. Two prominent settings for this modality include \textbf{rolling forecasting}~\citep{zhou2021informer, wu2021autoformer,zhang2023crossformer} and \textbf{temporal point process estimation}~\citep{mei2017neural,shchur2019intensity,
zuo2020transformer,xue2023easytpp}. \textit{Rolling forecasting} is commonly used for applications like weather forecasting or load monitoring. In this setting, observations are recorded at regular time intervals. At each time step $t$, one uses a lookback window of the past $L$ observations $\{ \mathbf{x}_{t-L}, \cdots \mathbf{x}_{t-1}\}$ to predict the next $H$ future steps $\{ \mathbf{x}_{t}, \cdots \mathbf{x}_{t+H-1}\}$~\citep{zhou2021informer}. \textit{Temporal point processes} are often used for tasks like behavior modeling and critical event forecasting. In this setting, one is interested in modeling discrete events that occur at potentially irregular intervals in continuous time. Given past events $\{ (t_1, c_1), \cdots (t_{n-1}, c_{n-1})\}$, where $t_i$ denotes the time and $c_i$ the type of event $i$, the goal is to predict the arrival time and type of the next event~\citep{xue2023easytpp}. 
Our new \textbf{arXiv} and \textbf{GitHub} datasets are relevant for such settings: the dense lead channels can be used for rolling forecasting, while the sparse lag channels can be reframed as temporal point processes. 
Beyond this contribution to the broader time-series community,
though, our proposed lead-lag forecasting task differs from both settings. Unlike rolling forecasting, which often predicts near-future outcomes immediately following the context window, we introduce a gap between the lead window and the prediction horizon.
Unlike temporal point processes, the lead channels in lead-lag prediction can consist of dense, regularly sampled measurements available at each time step.
Our datasets introduce exciting challenges for time series prediction: models must generalize to unseen series, transfer information across channels, and handle substantially longer prediction ranges than existing benchmarks. These open new opportunities for developing scalable, flexible forecasting architectures.\footnote{A detailed discussion of these challenges appears in the Supplementary Material, which can be accessed using the link to the arXiv preprint on our project website (\url{https://lead-lag-forecasting.github.io}).}

\textbf{Popularity and Conversion Prediction.}
Beyond general forecasting, specific attention has been given to predicting the future impact or conversion of items based on early user interactions.
In the scientific domain, \citet{brody2006earlier} and \citet{haque2009positional} pioneered the analysis of early web usage statistics, demonstrating that download frequencies in the first weeks of a preprint's life are strong predictors of later citation impact.
In the social media domain, popularity prediction often relies on modeling information cascades or self-exciting processes to forecast future shares or views~\citep{cheng2014can, zhao2015seismic}.
Similarly, in e-commerce, methods like ESMM (Entire Space Multi-task Model)~\citep{ma2018entire} leverage sequential user behaviors (e.g., click $\to$ purchase) to improve conversion rate prediction by modeling the conditional probability of delayed feedback.
While these works share the spirit of using early signals to predict later outcomes, they typically focus on short-term user-specific actions or single-metric extrapolation.
Our work generalizes these distinct problem settings into a unified cross-channel framework, explicitly targeting long-horizon prediction from short-term lead signals across entirely new series.

\textbf{Time Series Datasets.} We provide a summary and comparison of existing time series datasets in ~\autoref{tab:datasets}. These datasets are mostly designed for the rolling forecasting and temporal point process tasks discussed above. Popular datasets for rolling forecasting cover domains like energy and weather
\citep{electricityloaddiagrams20112014_321, nrel2025solar, bgcjena2025weather, Lai_2018, iif2025timeseries}. Common datasets for temporal point processes encompass domains like transportation and online activity
\citep{nyctlc2025tripdata, mo2023ubernyc, caltrans2025pems, stolfo1999kddcup, muonneutrino2017wikitraffic, zhao2015seismic}.
While these datasets have driven advancement in their respective areas, they do not capture clear lead-lag relationships central to our task---they omit two axes that are \emph{central} to LLF:
(i) \emph{Cross-channel prediction:} models must map an early ``lead'' channel to a delayed ``lag'' channel related to the same entity. 
(ii) \emph{Cross-series generalization:} models are trained on many entities and evaluated on \emph{new} entities that appear only at test time.

Although there are lines of work developing cross-series/variate capability with these datasets \citep{zhou2021informer}, primary benchmark tasks still focus on same-variate prediction.
Furthermore, the methods are not geared toward learning from thousands---or even millions---of series in order to forecast the future for hundreds of unseen ones.
For example, traffic datasets treat each sensor (e.g., a camera on a highway) as a variable and extrapolate its own future load; they rarely ask the model to forecast demand at a brand-new sensor from patterns observed at other sensors.

In contrast, \textbf{\arxiv} provides training data on over 1M papers’ access and citation trajectories and predicts citations for newly submitted papers given only an early access window, while \textbf{\github} provides data on pushes \& stars of 938K existing repositories to forecast forks of repositories that are unseen during training.  Both datasets are an order of magnitude larger than popular rolling-forecasting corpora and introduce research domains—scholarly communication and open-source software—that are absent from prior work. They therefore constitute the first large-scale benchmarks that \emph{jointly} test cross-channel and cross-series generalization, filling a critical gap in the time-series literature.

\begin{figure*}[t]

    \includegraphics[width=.7\linewidth]{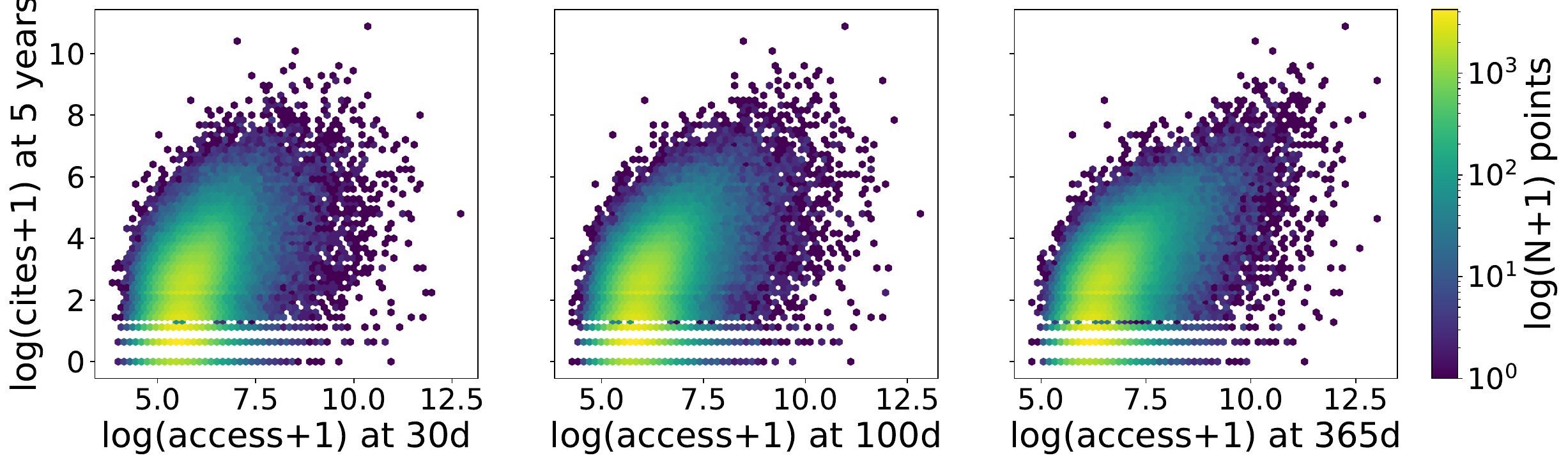}

    \caption{Hexbin plots of log-transformed early accesses vs. five-year citations. Each subplot corresponds to the 30-, 100-, and 365-day access horizon. Color indicates the log-density of papers. These plots illustrate a clear positive association across all horizons, with the signal becoming sharper and more linear at longer horizons.
    }
    \label{fig:arxiv-cor}
    \vspace{-3 mm}
\end{figure*}
\section{Problem Setting}
\label{sec:problem}


Consider an entity, e.g., an academic paper or a code repository, \(i\in\{1,\dots,N\}\) observed on a platform that logs data containing \emph{lead} and \emph{lag} channels. The lead and lag sequences are discrete-time signals sampled at a predefined interval (e.g., daily or weekly) starting when the entity initializes (e.g., a preprint is posted). Early in the timeline, the lag channel may be sparse or entirely zero, as measurable impact often takes time to manifest.
In particular, we define \textbf{lead channel} as a multivariate sequence \(\mathbf{x}^{(i)}_{1:T_i}\!=\!(x^{(i)}_{1},\dots,x^{(i)}_{T_i})\in\mathbb{R}^{T_i\times d_x}\) which captures early-phase interactions (e.g.,\ downloads, pushes, views).
The
\textbf{lag channel} is  multivariate sequence  
\(\mathbf{y}^{(i)}_{1:T_i} = (y^{(i)}_1, \dots, y^{(i)}_{T_i}) \in \mathbb{R}^{T_i \times d_y}\)
recording delayed signals of impact (e.g., citations, sales), aligned to the same absolute timeline.
We fix a prediction horizon \(H \in \mathbb{Z}_{>0}\), and define a \emph{lookback window} \(\tau < H\).  
We are interested in making a prediction  based only on early observation up to cutoff \(\tau\) about some behavior of \(\mathbf{y}^{(i)}\) at a much later timepoint \(H\).
In full generality, the learner may observe both early lead and lag sequences \(\mathbf{x}^{(i)}_{1:\tau},\mathbf{y}^{(i)}_{1:\tau}\), allowing models to leverage auto-regressive signals alongside cross-channel indicators, though the early lag sequence may be sparse or uninformative.
At prediction time, there is
no access to the future lead values \(\mathbf{x}^{(i)}_{\tau+1:H}\) nor lag values \(\mathbf{y}^{(i)}_{\tau+1:H}\).  
The gap between $\tau$ and $H$ may be significant, and this sets the LLF setting apart from much related work on time series 
forecasting.
The goal is to predict a downstream outcome derived from the lag channel at \(H\):  
\[
    \hat{z}^{(i)} = f_\theta\!\bigl(\mathbf{x}^{(i)}_{1:\tau}, \mathbf{y}^{(i)}_{1:\tau}\bigr)
    \quad\text{with target}\quad
    z^{(i)} = g\bigl(\mathbf{y}^{(i)}_{1:H}\bigr),
\]
where \(g(\cdot)\) extracts a quantity of interest, like the value at a fixed future timepoint (e.g., \(y^{(i)}_{H}\)), a cumulative sum, or a binary label (e.g., top‑10\% impact).
For example, we may use a sequence of preprint accesses over the first $\tau=30$ days to predict whether the paper will receive at least 50 citations in the next $H=5$ years.
Unlike rolling forecasting~\citep{zhou2021informer}, LLF  
(i) ignores the intermediate values between \(\tau\) and \(H\);
(ii) leverages cross-channel dependence between \(\mathbf{x}\) and \(\mathbf{y}\); and  
(iii) focuses on horizons where \(H\) is large relative to typical seasonality patterns, making naïve extrapolation ineffective.

We provide and analyze two novel benchmark datasets for investigating LLF:
\begin{enumerate}
  \item \textbf{\arxiv}\,: accesses, look-back \(\mathcal{L}=\{30,100,365\}\) $\rightarrow$ citations, horizons \(\mathcal{H}=\{1825\}\) days;
  \item \textbf{\github}\,: pushes, stars $\rightarrow$ forks, same \(\mathcal{L}\) and \(\mathcal{H}\).
\end{enumerate}
We fix a long-term forecast horizon \(H = 1825\) days (5 years) across both datasets for two primary reasons: first, five years is a widely accepted benchmark in bibliometrics and open-source software for evaluating sustained impact (e.g., the ``C5'' metric~\citep{yu2014citation}); second, a five-year horizon presents a non-trivial forecasting challenge where simple short-term extrapolation often fails, necessitating models that learn underlying lead-lag dynamics. Meanwhile, extending beyond five years would significantly reduce the volume of training data available without survivorship bias.

Rather than varying the prediction target, we vary the \emph{lookback window}: the length \(\tau\) of the available early lead signal.
In Section~\ref{sec:arxiv} and~\ref{sec:github} we investigate whether early-phase signals \(\mathbf{x}^{(i)}_{1:\tau}\) and \(\mathbf{y}^{(i)}_{1:\tau}\), are predictive of a lag outcome \(z^{(i)} = g(\mathbf{y}^{(i)}_{\tau+1825})\).  
This setup enables us to study how much predictive power accumulates as more early interaction data becomes available.

We additionally outline a standard methodology for evaluating  predictions, and present accuracy results for several baseline supervised learning models in Section~\ref{sec:experiments} across \(\tau\in\mathcal{T} = \{30, 100, 365\}\) days.
Many lead-lag signals of interest follow a heavy tailed, power law distribution, and therefore it's important to use appropriate metrics to measure performance.
Given a labelled dataset of
$N_{\text{test}}$ examples for evaluating performance,
we consider metrics for two types of tasks:
    \textbf{(1) Classification}, which aims to identify high-impact entities defined by a threshold binarization of the lag outcome. Here, the metrics are AUROC and F1 score: the AUROC captures the model's ability to rank positive examples higher than negative ones across all thresholds, while the F1 score summarizes the precision and recall at a fixed decision threshold (e.g., predicting whether a paper will receive more than 50 citations).
    \textbf{(2) Regression}, which aims to predict the numerical values of the lag outcomes. The three evaluation metrics are mean absolute error in the linear (\textbf{MAE}) and log (\textbf{MAE-log}) space (computed with $z\leftarrow \log(1+z)$), and mean absolute percentage error (\textbf{MAPE}) in the linear space for high-impact entities ($z\geq k$). 
\begin{equation}\label{eq:mae}
\begin{aligned}
    \text{MAE} &= \frac{1}{N_{\text{test}}} \sum_{i=1}^{N_{\text{test}}} \big\vert \hat{z}^{(i)} - z^{(i)} \big\vert, \\
\text{MAPE} &= \frac{1}{|\mathcal{I}|} \sum_{i \in \mathcal{I}} \left| \frac{\hat{z}^{(i)} - z^{(i)}}{z^{(i)}} \right| ~\text{for}~\mathcal{I} = \left\{ i \mid z^{(i)} \geq k \right\}
\end{aligned}
\end{equation}
MAE is a direct measure of the average prediction from the ground truth outcomes, while MAE-log accounts for the long-tailed nature. MAPE quantifies the relative errors between the predicted and ground truth outcomes. We restrict this metric to high impact entities with $z\geq k$, to avoid instability from low-impact instances, e.g. $\hat z = 2$ when $z=1$ yields 100\% error but is inconsequential.





\section{\arxiv Dataset}\label{sec:arxiv}
\begin{figure*}[t]
    \includegraphics[width=.7\linewidth, trim={0 0 0 0.2cm}, clip]        {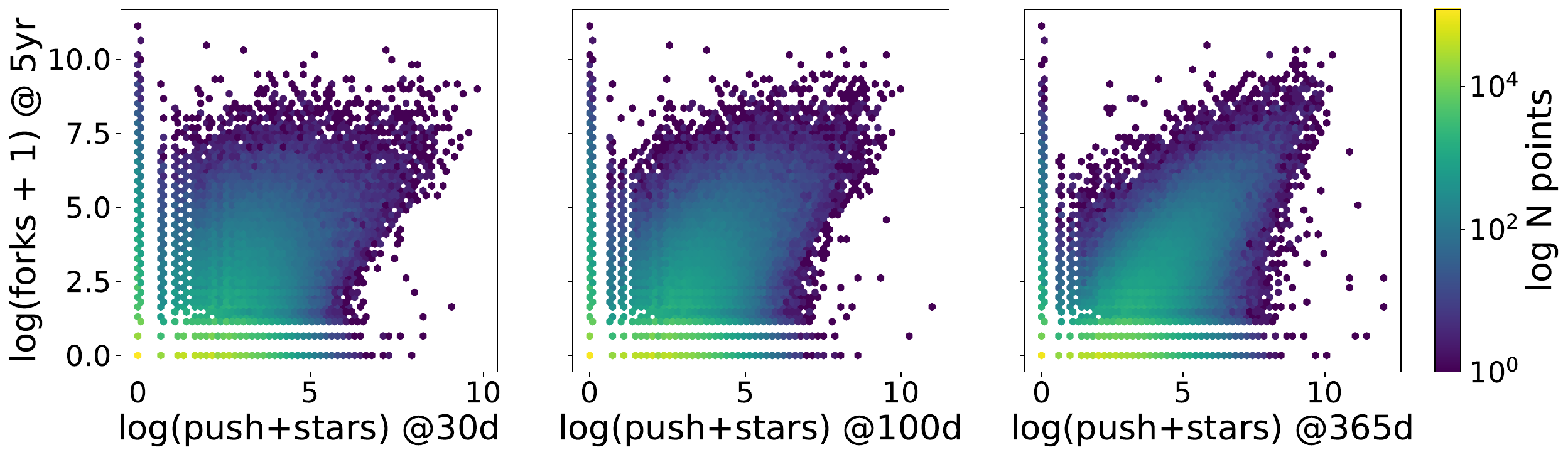}
    \caption{Hexbin plots of log-transformed early push+star events vs. five-year forks. Each subplot corresponds to the 30-, 100-, and 365-day horizons. Color indicates the log-density of repositories. These plots illustrate a positive association across all horizons, with the signal becoming sharper and more linear at longer horizons. }
    \label{fig:github_hexbin}
\end{figure*}

\subsection{Processing, Statistics, and Distribution}


\textbf{Data Sources.}
The dataset is built by a \emph{left join} between
\textbf{(i) the arXiv access logs} and
\textbf{(ii) the citation graph in the Semantic Scholar} JSON dump
extracted via the Datasets API \footnote{Semantic Scholar Datasets API: \url{https://api.semanticscholar.org/api-docs/datasets}.}. 

In \textit{\arxiv Access Logs}, the raw file records 2.3M
arXiv preprint IDs together with their submission dates and 4.87B access events from 1.07B anonymized users during the time period.
Coverage is weekly from \texttt{2006-07-03} to \texttt{2013-06-30},
and daily from \texttt{2013-07-01} onward.  
We aggregate the access times into a (potentially sparse) access time series for each paper. In \textit{Semantic Scholar Citations}, the dump contains 220M paper metadata records, from which we extracted 2.66B
directed citation pairs $(\textit{citing},\textit{cited})$.
For any pair in which the \emph{cited} paper has an arXiv ID,
we approximate the citation time by the publication date of the
\textit{citing} paper (true citation dates are unavailable) and
append it to that paper’s citation sequence.  
This yields citation time series for 2M
arXiv papers.

\textbf{Merging and Anonymization.}
Performing a left join between access and citation data produces 2.3M unique papers with recorded arXiv submission,
of which 2M 
have \emph{both} citation and access records.
For anonymity we drop paper IDs and submission dates in the released files.
Feature descriptions appear in the Appendix.

\textbf{Data Splits.}
Since \arxiv switched from weekly to daily access logs on \texttt{2013-07-01},
and a small fraction (1.88\%) of papers have other anomalies (see Appendix),
we set aside these cases in a \texttt{train\_extra} split.
We then sample random \texttt{train} / \texttt{val} / \texttt{labeled\_test}
splits from papers published between
\texttt{2013-07-01}
and \texttt{2018-09-25}.
The distribution of paper publication dates is illustrated in Figure~\ref{fig:coarse_category} in Appendix~\ref{app:arxiv-details} (left).
Papers submitted after \texttt{2018-09-25} form an \texttt{unlabeled\_test} set. The final split sizes are 318K, 111K, 79K, 681K, and 1.14M for \texttt{train}, \texttt{val}, \texttt{labeled\_test}, \texttt{unlabeled\_test}, and \texttt{train\_extra} respectively.

\textbf{Subject-area Distribution.}
There are 152 fine-grained and 20 coarse first-choice categories.
\autoref{fig:coarse_category} in Appendix~\ref{app:arxiv-details} (right) plots the coarse distribution for the train set.
The four largest areas are \verb|cs|, \verb|math|, \verb|cond-mat|, and \verb|astro-ph|.
Randomised splitting preserves this mix, whereas the \texttt{unlabeled test}
split---being more recent---contains a noticeably higher share of \verb|cs| papers.



\begin{figure}[t]
    \centering

    \includegraphics[width=\linewidth]
    {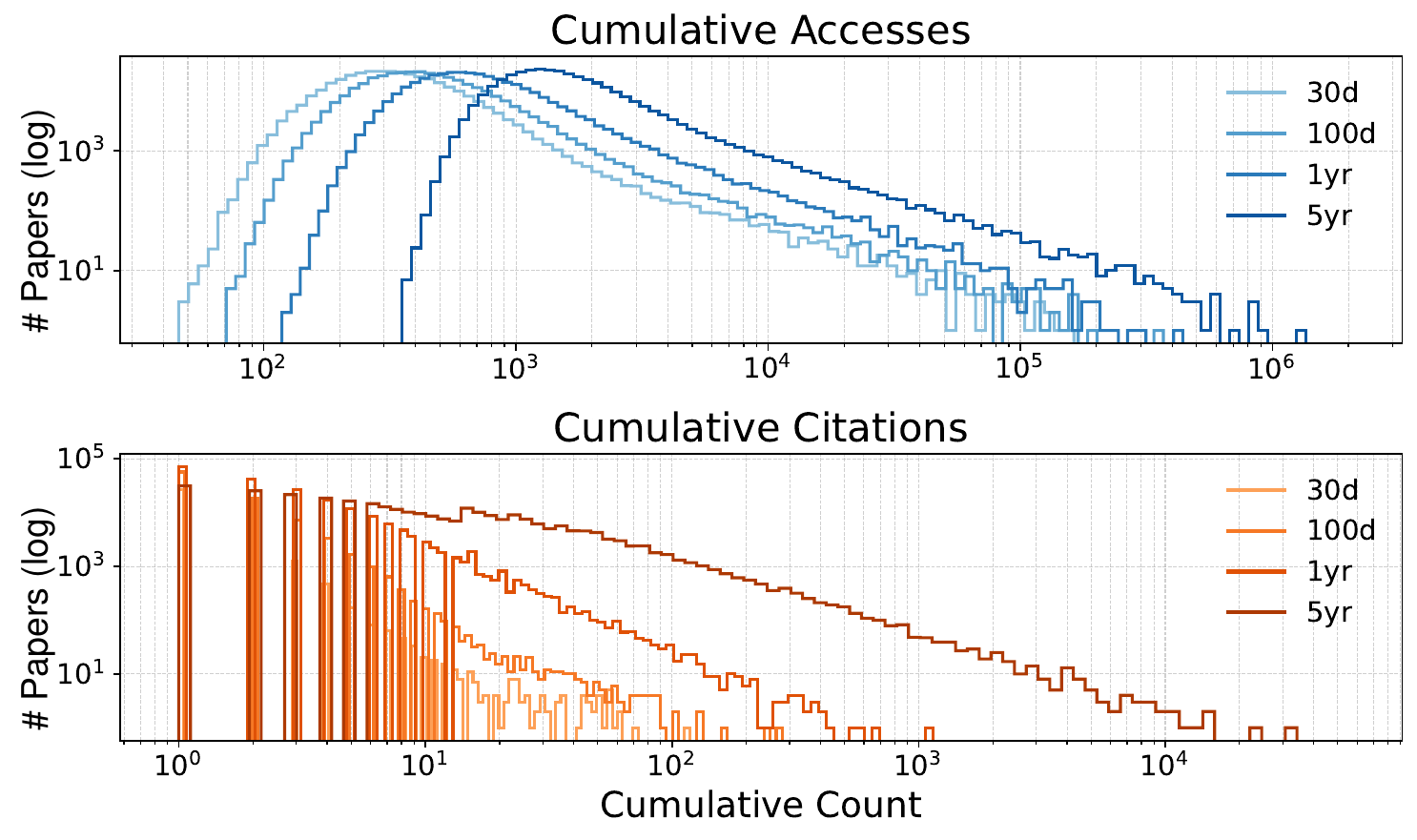}

    \caption{Distribution of accesses and citations over time. Log-scaled histograms show cumulative accesses (top) and citations (bottom) at 30 days, 100 days, 1 year, and 5 years after preprint submission. Colors progress from light to dark over time. } 
    \label{fig:temporal_dynamics}
    \vspace{-6mm}
\end{figure}

\subsection{Lead and Lag Analysis}

We investigate the lead and lag dynamics in the \arxiv dataset to assess whether early access metrics provide reliable indicators of future scholarly recognition. Our analysis focuses on cumulative accesses and citations at 30 days, 100 days, and 5 years.

\textbf{Channel Dynamics.} \autoref{fig:temporal_dynamics} illustrates the temporal dynamics of accesses and citations, showing that accesses clearly \emph{lead} citations, which \emph{lag} in numbers. Across the corpus, the first access consistently occurs before the first citation. At later times (darker shades), both distributions shift rightward and citations become increasingly dispersed and heavy tailed, reflecting 
variation in long-term impact. Notably, accesses distributions are bell-shaped even on shorter timescales, suggesting more predictable access patterns. 
These observations support early accesses as potential predictors of long-term scholarly recognition.

\textbf{Predictive Power.} \autoref{fig:combined_analysis} (2) shows the Pearson correlation between early citations/accesses (cumulative counts) with 5-year citations. Early accesses exhibit stronger correlation with long-term impact than citations, which only surpass accesses around day 50. This highlights citation's delayed utility. \autoref{fig:arxiv-cor} presents log-log hexbin plots of early accesses vs. long-term citations, revealing a clear, increasingly linear association over time. Density patterns also reveal a floor effect, where some papers receive early attention but few citations. These findings support early usage metrics as effective early indicators of scientific impact.

\begin{table*}[htbp]
\centering
\scriptsize
\caption{\textbf{Classification results for arXiv citation prediction across different horizons.}}
\vspace{-2mm}
\label{tab:arxiv_performance}
\resizebox{0.9\textwidth}{!}
{%
\begin{tabular}{llccccccccc}
\toprule
\multirow{2}{*}{\textbf{Horizon}} & \multirow{2}{*}{\textbf{Method}}
& \multicolumn{3}{c}{\textbf{Citations from Cit.}}
& \multicolumn{3}{c}{\textbf{Citations from Acc.}}
& \multicolumn{3}{c}{\textbf{Citations from (Cit.+Acc.)}} \\
\cmidrule(lr){3-5} \cmidrule(lr){6-8} \cmidrule(lr){9-11}
& & F1 & AUC & AUC-PR & F1 & AUC & AUC-PR & F1 & AUC & AUC-PR \\
\midrule
& Stratified Random      & 0.082 & 0.50 & 0.082  & 0.082 & 0.50 & 0.082  & 0.082 & 0.50 & 0.082  \\
& All-positive & 0.151 & 0.50 & 0.082  & 0.151 & 0.50 & 0.082  & 0.151 & 0.50 & 0.082  \\
\midrule
\multirow{4}{*}{\textbf{30 Days}}
& One-feature LR & 0.271 & 0.618 & 0.164 & 0.313 & 0.800 & 0.317 & 0.334 & 0.817 & 0.361 \\
& Vanilla LR      & \textbf{0.281} & \textbf{0.622} & 0.180 & \textbf{0.352} & 0.824 & 0.331 & 0.369 & 0.842 & 0.371 \\
& Time-MoE LR    & \textbf{0.281} & \textbf{0.622} & \textbf{0.183} & 0.349 & 0.836 & 0.390 & \textbf{0.377} & 0.846 & 0.406 \\
& InceptionTime    & 0.068 & \textbf{0.622} & 0.181 & 0.259 & \textbf{0.840} & \textbf{0.401}  & 0.306 & \textbf{0.855} & \textbf{0.438} \\
\midrule
\multirow{4}{*}{\textbf{100 Days}}
& One-feature LR & 0.299 & \textbf{0.756} & 0.318 & 0.348 & 0.832 & 0.366 & 0.398 & 0.870 & 0.475 \\
& Vanilla LR      & 0.304 & 0.755 & 0.351 & 0.368 & 0.851 & 0.399 & 0.434 & 0.889 & 0.503 \\
& Time-MoE LR    & \textbf{0.334} & 0.753 & \textbf{0.360} & \textbf{0.425} & 0.866 & 0.466 & 0.425 & 0.895 & 0.533 \\
& InceptionTime    &  0.265 & \textbf{0.756} & 0.358 &  0.359 & \textbf{0.867} & \textbf{0.480} & \textbf{0.462} & \textbf{0.901} & \textbf{0.574}\\
\midrule
\multirow{4}{*}{\textbf{365 Days}}
& One-feature LR & 0.462 & 0.929 & 0.666 & 0.388 & 0.865 & 0.433 & 0.534 & 0.948 & 0.714 \\
& Vanilla LR      & 0.485 & \textbf{0.932} & 0.712 & \textbf{0.397} & 0.870 & 0.453 & 0.562 & 0.954 & 0.746 \\
& Time-MoE LR    & 0.508 & 0.929 & 0.700 & 0.388 & \textbf{0.882} & 0.506 & 0.569 & 0.956 & 0.758 \\
& InceptionTime   & \textbf{0.618}	& \textbf{0.932} & 0\textbf{.717} & 0.381 &  \textbf{0.882} & \textbf{0.509} & \textbf{0.681} & \textbf{0.960}  & \textbf{0.781} \\
\bottomrule
\end{tabular}}
\end{table*}

\section{\github Dataset}\label{sec:github}

\subsection{Processing, Statistics, and Distribution}

\textbf{Data Sources.} The GitHub event data are from GH Archive and hosted on Google BigQuery~\citep{grigorik2025gharchive}, for which the coverage starts from \texttt{2011-02-12}. Events associated with 424M
repositories are recorded between \verb|2011-02-12| and \verb|2024-12-31|. We select pushes, creates (creating a repository, branch, or Git tag), stars, and forks as our primary focus.

Due to the sparse nature of GitHub events, we focus on those associated with packages that have been released on some manager platform (e.g. PyPI, Conda, \textit{etc.}). Specifically, we extract repository metadata associated to packages on 32 platforms recorded on Ecosyste.ms. We used an open-access data dump from Ecosyste.ms that covers 3M
package repositories created between \texttt{2007-10-29} and \texttt{2024-06-04} \footnote{Ecosyste.ms open data releases: \url{https://packages.ecosyste.ms/open-data}.}.
We restrict to the package repositories created within the coverage of GH Archive (\texttt{2011-02-12} to \texttt{2024-12-31}). The left joining of the Ecosyste.ms metadata with the GH Archive event data results in 3M
repositories in total, 2.9M
of which have associated events recorded in GH Archive.
The features of the resulting final dataset are detailed in the Appendix.

\textbf{Data Splits.} We generated a random split of sizes: 938K train, 235K test, and 235K validation. The repositories with histories shorter than 1825 days (5 years) go into the unlabeled test set of size 1.57M, and the examples with problematic first events are saved to an extra training set of size 58K. \autoref{fig:plaform-dist} in Appendix~\ref{app:github-details} shows the creation timeline of GitHub repositories in the dataset and the distribution of top 15 packages. Additional dataset details are included in the Appendix.

\subsection{Lead and Lag Analysis}
To investigate whether early engagement patterns can predict long-term repository impact on \github, we examine how different types of engagement signals on repositories evolve over time. Our analysis tracks cumulative counts of pushes, stars, and forks at days 30, 100, 365, and 5 years.

\begin{figure}[t]

    \includegraphics[width=\linewidth]{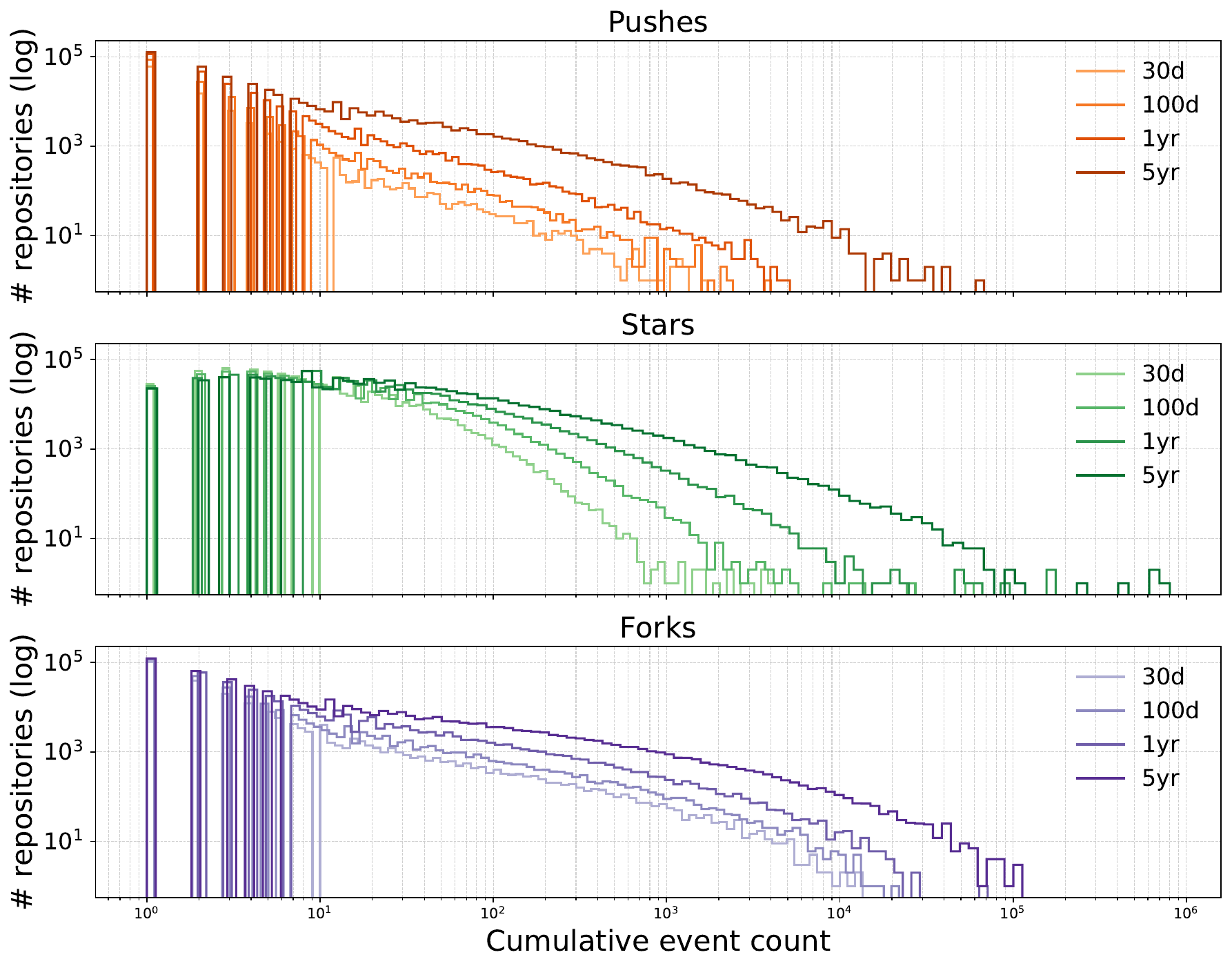}
    \caption{Distributions of GitHub engagement signals across time horizons. Log-scaled histograms showing cumulative counts of pushes (top), stars (middle), and forks (bottom) measured at 30 days, 100 days, 1 year, and 5 years after repository creation. Colors progress from lighter to darker as time advances.}
    \label{fig:github-arrival}
\end{figure}
\textbf{Channel Dynamics.} \autoref{fig:combined_analysis} (3) shows the first arrival time of events after the first push across repositories. The median repository receives its first star within approximately 10–30 days, whereas the first fork typically occurs after 100–200 days. By the 90th percentile, stars appear within a few hundred days, while forks may take several years. This temporal gap indicates that community engagement generally begins with a star and only later escalates to a fork. \autoref{fig:github-arrival} shows engagement signal distributions across channels and time horizons. As time progresses from 30 days to 5 years (lighter to darker lines), all distributions shift right and widen, reflecting increasing heterogeneity in repository popularity and activity. Forks display the most rightward spread among the three, highlighting their role as a signal of sustained engagement akin to citations. Therefore, our benchmarks use forks as the target, and pushes and stars as early indicators. We leave the exploration of alternative choices of lead-lag pairs as future work.



\textbf{Predictive Power.} \autoref{fig:combined_analysis} (4) shows their correlations with 5-year forks. Notably, early stars exhibit stronger correlation with long-term forks than early forks. \autoref{fig:github_hexbin} visualizes the relationship between early activity and long-term forks via log-log hexbin plots, revealing a consistent positive association that sharpens over time. Density patterns also show floor effects at low activity levels. These observations support early pushes and stars as practical early indicators of forks.

\begin{table*}[htbp]
\centering
\scriptsize
\caption{\textbf{Classification results for GitHub fork prediction across different horizons.}}
\label{tab:github_performance}
\resizebox{\textwidth}{!}
{%
\begin{tabular}{llccccccccccccccc}
\toprule
\multirow{2}{*}{\textbf{Horizon}} & \multirow{2}{*}{\textbf{Method}}
& \multicolumn{3}{c}{\textbf{Forks from Pushes}}
& \multicolumn{3}{c}{\textbf{Forks from Stars}}
& \multicolumn{3}{c}{\textbf{Forks from Forks}}
& \multicolumn{3}{c}{\textbf{Forks from (P+S)}}
& \multicolumn{3}{c}{\textbf{Forks from (P+S+F)}} \\
\cmidrule(lr){3-5} \cmidrule(lr){6-8} \cmidrule(lr){9-11} \cmidrule(lr){12-14} \cmidrule(lr){15-17}
& & F1 & AUC & AUC-PR & F1 & AUC & AUC-PR & F1 & AUC & AUC-PR & F1 & AUC & AUC-PR & F1 & AUC & AUC-PR \\
\midrule
& Stratified Random        & 0.113 & 0.50 & 0.113 & 0.113 & 0.50 & 0.113 & 0.113 & 0.50 & 0.113 & 0.113 & 0.50 & 0.113 & 0.113 & 0.50 & 0.113 \\
& All-positive & 0.204 & 0.50 & 0.113 & 0.204 & 0.50 & 0.113 & 0.204 & 0.50 & 0.113 & 0.204 & 0.50 & 0.113 & 0.204 & 0.50 & 0.113 \\
\midrule
\multirow{4}{*}{\textbf{10 Days}}
& One-feature LR & 0.189 & 0.534 & 0.140 & 0.296 & 0.639 & 0.254 & 0.281 & 0.592 & 0.212 & 0.294 & 0.679 & 0.283 & 0.308 & 0.695 & 0.306 \\
& Vanilla LR      & 0.214 & 0.596 & 0.156 & 0.306 & 0.640 & 0.277 & \textbf{0.283} & 0.591 & 0.238 & \textbf{0.308} & 0.698 & 0.306 & \textbf{0.327} & 0.713 & 0.333 \\
& Time-MoE LR    & \textbf{0.245} & 0.623 & \textbf{0.170} & \textbf{0.325} & 0.641 & \textbf{0.286} & \textbf{0.283} & 0.591 & \textbf{0.239} & 0.294 & 0.725 & \textbf{0.334} & 0.304 & 0.722 & \textbf{0.348} \\
& InceptionTime   & 0.000	& \textbf{0.670}	& 0.064 &	0.103	& \textbf{0.664}	& 0.176	& 0.127	& \textbf{0.622}	& 0.167	& 0.109	& \textbf{0.772}	& 0.196	& 0.151	& \textbf{0.780}	& 0.220  \\
\midrule
\multirow{4}{*}{\textbf{30 Days}}
& One-feature LR & 0.210 & 0.578 & 0.169 & 0.328 & 0.694 & 0.332 & 0.369 & 0.651 & 0.300 & 0.335 & 0.727 & 0.359 & 0.369 & 0.749 & 0.399 \\
& Vanilla LR      & 0.249 & 0.631 & 0.187 & 0.369 & 0.696 & 0.360 & \textbf{0.376} & 0.650 & \textbf{0.334} & \textbf{0.373} & 0.751 & 0.388 & \textbf{0.397} & 0.769 & 0.424 \\
& Time-MoE LR    & \textbf{0.261} & 0.668 & \textbf{0.211} & \textbf{0.378} & 0.689 & \textbf{0.387} & \textbf{0.376} & 0.661 & \textbf{0.334} & 0.314 & 0.768 & \textbf{0.420} & 0.339 & 0.777 & \textbf{0.447} \\
& InceptionTime   & 0.000	& \textbf{0.715}	& 0.088	& 0.243	& \textbf{0.736}	& 0.285	& 0.249 & 	\textbf{0.697}	&  0.279	& 0.226	& \textbf{0.830}	& 0.302	& 0.272 & 	\textbf{0.839}	&  0.338 \\
\midrule
\multirow{4}{*}{\textbf{100 Days}}
& One-feature LR & 0.248 & 0.640 & 0.216 & 0.435 & 0.773 & 0.454 & 0.456 & 0.756 & 0.466 & 0.426 & 0.800 & 0.473 & 0.466 & 0.831 & 0.547 \\
& Vanilla LR      & \textbf{0.294} & 0.651 & 0.231 & 0.432 & 0.778 & 0.491 & 0.469 & 0.755 & 0.511 & 0.449 & 0.825 & 0.507 & 0.488 & 0.849 & 0.580 \\
& Time-MoE LR    & 0.282 & 0.717 & \textbf{0.265} & \textbf{0.468} & 0.777 & \textbf{0.532} & \textbf{0.482} & 0.746 & \textbf{0.522} & \textbf{0.487} & 0.844 & \textbf{0.556} & \textbf{0.505} & 0.864 & \textbf{0.615} \\
& InceptionTime    & 0.002 & 	\textbf{0.778}	& 0.137	& 0.429	& \textbf{0.824}	& 0.460	& 0.453	& \textbf{0.805}	& 0.474	& 0.432	& \textbf{0.898}	& 0.474	& 0.476 & 	\textbf{0.908} & 	0.525 \\
\bottomrule
\end{tabular}}
\end{table*}
\vspace{0.8\baselineskip}
\section{Baseline Experiments}
\vspace{0.8\baselineskip}
\label{sec:experiments}
We now present results for lead-lag prediction using baseline supervised learning methods.
Our goal is twofold: 
to empirically validate the presence of predictive signal in our datasets,
and to establish baseline performance benchmarks and evaluation methodology to ground future research.\par
\vspace{0.5\baselineskip}

\textbf{High-Impact Lag Classification.}
\label{sec:classification}
We first explore whether early activity alone can predict which items will land in the top $8.1$\% of \arxiv\ papers ($\ge50$ citations) or top $11.3$\% of \github\ repos ($\ge10$ forks) after five years.
We perform classification with logistic regression in three settings: 
\begin{enumerate}
    \item using one feature---the cumulative count of the lead signal at time $\tau$;
    \item using the raw features up to time $\tau$;
    \item using a time-series foundation model---Time-MoE (large)~\citep{shi2024time} embeddings of raw features with a logistic regression head. We perform ablations across all channels, and pairs of channels, for both dataset and all model.
\end{enumerate}
Additionally, we consider a representative deep learning-based time-series classification method, InceptionTime~\citep{ismail2020inceptiontime}.\footnote{Model descriptions and implementation details for the baseline experiments are presented in the Supplementary Material, which can be accessed using the link to the arXiv preprint on our project website (\url{https://lead-lag-forecasting.github.io}).} 

\begin{figure*}[t]
    \includegraphics[width=\linewidth]{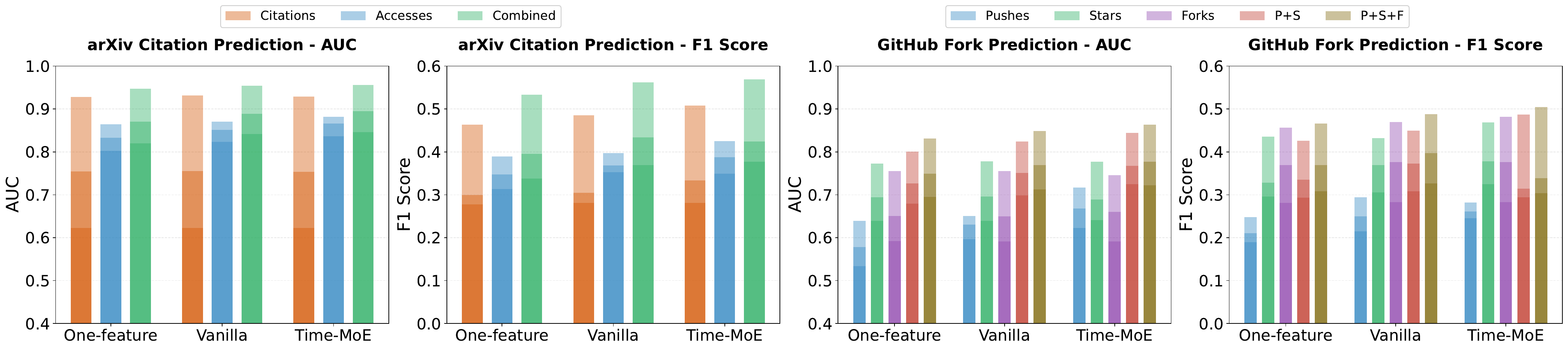}
    \caption{Classification results on \arxiv (left) and \github (right) across different channels, models and input-horizon. Cross-channel prediction outperforms same-channel prediction for early input-horizon, consistent with correlations plots in \autoref{fig:github_hexbin} and \autoref{fig:arxiv-cor}. Deep features generally outperform non-transformed features, while even classification with a single feature outperforms random (F1 0.08 for \arxiv and F1 0.11 for github). As we increase the input horizon models perform better.}
    \label{fig:lr-results}
\end{figure*}

The results in \autoref{tab:arxiv_performance} and \autoref{tab:github_performance}, illustrated in \autoref{fig:lr-results}, demonstrate several observations; 
\begin{enumerate}
    \item Predictive power even with one feature: using the single cumulative feature, the AUC climbs from 0.80 (day 30) to 0.86 (day 365) while F1 rises from 0.31 to 0.39 for \arxiv from cross-channel prediction, and outperform same-channel predictions up to day 100, consistent with our correlation results in \autoref{fig:arxiv-cor}. GitHub exhibits a similar pattern consistent with \autoref{fig:github_hexbin}.
    \item Cross-channel prediction outperforms same-channel prediction for early input-horizon, across all models. This is consistent with correlations plots in \autoref{fig:github_hexbin} and \autoref{fig:arxiv-cor}
    \item Using more channels (e.g. pushes + stars) performs better than single channel prediction across all models. 
    \item Using all features generally outperforms using a single feature, and Deep features(Time-MoE embeddings) generally outperform non-transformed features in most cases, indicating that more expressive features are potentially helpful.
    \item In all cases, predictive performance exceeds that of random baselines, providing clear evidence that early usage patterns contain meaningful signal about long-term impact.
    \item InceptionTime demonstrates superior performance under long prediction horizon and ranking-based metric (AUC), but demonstrates lower F1 for shorter horizons possibly due to miscalibration. 
\end{enumerate}

\textbf{Lag Outcome Regression.}
We next provide baseline benchmarks in the regression setting.
We evaluate models using metrics defined in \autoref{sec:problem}. For MAPE, we use threshold $k=50$ for \arxiv and $k=10$ for \github to designate high-impact items. We consider the standard application of five methods: MEY~\citep{abrishami2019predicting}, Linear Regression, KNN, MLP, Transformer~\citep{vaswani2017attention} and KNN with embeddings from Time-MoE (large)~\citep{shi2024time}. We train on the training split of each dataset, and evaluate on the test split. We use all channels as inputs for all baselines except MEY, and use only the lag channel for MEY, since MEY does not support cross-channel prediction.

We report the results on \arxiv in \autoref{tab:arxiv-benchmark} and results on \github in~\autoref{tab:github-benchmark}. In both cases, all methods generally achieve better performance as the input horizon increases from 30 days to 1 year, validating that more information facilitates prediction. Straightforward applications of deep learning methods yield performance gains, highlighting the potential for future research on more advanced deep learning approaches for this task. Notably, KNN attains strong performance, particularly on \github, for MAE metrics, but underperforms the deep learning methods on MAPE.
We hypothesize that this effect arises due to the skewed data distribution: on \github, many repos have zero forks in 5 years.
Deep learning methods show a clear advantage in predicting the high-impact instances.
KNN using Time-MoE embeddings achieves comparable results to standard KNN on \arxiv and \github.
Thus, while we see that the extracted features contain some signal, we do not find compelling evidence that the time-series foundation model adds substantial value for this task.

\begin{table*}[ht]
\centering
\scriptsize
\caption{\textbf{Benchmarking results on the test split of the Arxiv dataset.} 
}
\vspace{-2mm}
\resizebox{.8\textwidth}{!}
{%
\begin{tabular}{lccc ccc ccc}
\toprule
\multirow{2}{*}{\textbf{Method}} 
& \multicolumn{3}{c}{\textbf{30 Days}} 
& \multicolumn{3}{c}{\textbf{100 Days}} 
& \multicolumn{3}{c}{\textbf{365 Days}} \\
\cmidrule(lr){2-4} \cmidrule(lr){5-7} \cmidrule(lr){8-10}
& MAE-log & MAE & MAPE
& MAE-log & MAE & MAPE
& MAE-log & MAE & MAPE \\
\midrule
MEY                & 2.057 & 24.902 & 0.934 & 1.631 & 19.770 & 0.758 & 0.835 & 13.534 & 0.516\\
Linear Regression  & 0.841 & 22.895 & 1.393 & 0.767 & 16.725 & 0.756 & 0.574 & 11.247 & 0.463\\
KNN                & 0.932 & 19.121 & \textbf{0.737} & 0.856 & 17.613 & 0.690 & 0.656 & 12.967 & 0.512 \\
MLP                & 0.842 & 17.651 & 0.763 & 0.771 & 16.193 & 0.678 & 0.576 & 11.656 & 0.459\\
Transformer        & \textbf{0.832} & \textbf{17.422} & 0.747 & \textbf{0.759} & \textbf{15.669} & \textbf{0.668} & \textbf{0.568} & \textbf{10.987} & \textbf{0.443}\\
Time-MoE KNN       & 0.931 & 18.954 & 0.732 & 0.857 & 17.613 & 0.686 & 0.657 & 13.232 & 0.521\\
\bottomrule
\end{tabular}}
\label{tab:arxiv-benchmark}

\end{table*}
\begin{table*}[t]
\centering
\scriptsize
\caption{\textbf{Benchmarking results on the test split of the \github dataset.}}
\vspace{-2mm}
\resizebox{.8\textwidth}{!}
{%
\begin{tabular}{lccc ccc ccc}
\toprule
\multirow[c]{2}{*}{\textbf{Method}} & \multicolumn{3}{c}{\textbf{30 Days}} & \multicolumn{3}{c}{\textbf{100 Days}} & \multicolumn{3}{c}{\textbf{365 Days}} \\
\cmidrule(lr){2-4} \cmidrule(lr){5-7} \cmidrule(lr){8-10}
 & MAE-log & MAE & MAPE & MAE-log & MAE & MAPE & MAE-log & MAE & MAPE \\
\midrule
MEY & 0.775	& 25.467	& 2.526	& 	0.649	& 15.721	& 1.464	& 	0.423	& 9.438 & 0.834\\
Linear Regression &  0.757	& 24.068	& 1.076		& 0.647	& 34.474	& 1.001 & 0.425 &	13.385	& 0.660 \\
KNN & \textbf{0.675}	& 11.882	& 0.923		& \textbf{0.556}	& 10.934 & 	0.875	& 	0.360	& 8.689 & 	0.672\\
MLP & 0.739 & 	11.834	& 0.868 & 0.622	& 10.814	& 0.800 & 0.408 &	8.664	& 0.601\\
Transformer & 0.732	& \textbf{11.456} & \textbf{0.856} & 0.612& \textbf{10.085}& \textbf{0.767} & \textbf{0.398}	& \textbf{7.530}	& \textbf{0.543}\\
Time-MoE KNN      & 0.716 & 12.05 & 0.925 & 0.578 & 11.11 & 0.872 & 0.355 & 8.92 & 0.686\\
\bottomrule
\end{tabular}}
\label{tab:github-benchmark}

\end{table*}

\label{sec:benchmarks}
\section{Discussion and Conclusion}
\label{sec:conclusion}
We establish Lead-Lag Forecasting (LLF) as
a formal prediction problem, motivated by the gap between observed \emph{lead-lag dynamics} in important domains---including scientific and technological impact---and popular time series forecasting benchmarks. 
We catalyze research on LLF by curating and releasing two novel datasets: \arxiv papers and \github repositories.
We establish lead-lag relationships in streams of activity data and provide baseline numbers for several standard supervised machine learning methods on the task of predicting a 5-year outcome from as little as one month of observation.
While our results demonstrate the existence of predictive signal, we speculate that there are opportunities for innovation to improve predictions.

\textbf{Limitations.} Our datasets contain only aggregate signals resulting from user activities, such as accesses, stars, and citations. 
They do not contain behavior traces at the individual level (i.e. \emph{who} downloaded which paper), nor do they contain additional information about entities, such as author lists or content.
It is likely that such information could provide complementary predictive power.
Furthermore, the datasets that we curate reflect the social dynamics of particular platforms at particular times, and it is plausible that such dynamics could change or be altered by interventions based on these early signals. Additionally, LLF data and models capture predictive correlations rather than causal relationships, and therefore should not be used to infer direct causal interventions. There are also limitations due to lag signals (citations, forks) serving only as approximations of quality, success, and impact.
Nevertheless, we hope these datasets will enable better understanding of these promises and pitfalls of predicting long term outcomes.

\textbf{Extension to Other Domains.} Beyond scientific and open-source ecosystems, the LLF framework naturally extends to a wide range of platforms exhibiting analogous dual-process dynamics. Motivating examples include Wikipedia (page-views → edits), Spotify (streams → concert attendance), e-commerce (click-throughs → purchases), and professional networks like LinkedIn (profile views → messages). While these domains are highly relevant, their coupled interaction data remains strictly proprietary, making our public release of arXiv and GitHub data uniquely valuable for open research. Furthermore, whereas commercial platforms are often dominated by fast-paced, algorithmically driven viral loops, the multi-year feedback cycles in our selected technical ecosystems offer a stable environment to model deep structural phenomena. We hope that establishing rigorous LLF baselines on these mathematically rich, long-horizon datasets will provide the theoretical and empirical foundation necessary for the community to eventually transfer these methodologies to noisier, high-velocity domains.


\begin{acks}
This work was funded by NSF CCF-2312774, NSF OAC-2311521, NSF IIS-2442137,
NSF TIP-2404035, NASA 80NSSC25M7039, a PCCW Affinito-Stewart Award, a gift
to the LinkedIn-Cornell Bowers CIS Strategic Partnership, an AI2050 Early
Career Fellowship program at Schmidt Sciences, and NewYork-Presbyterian
for the NYP-Cornell Cardiovascular AI Collaboration.
\end{acks}

\bibliographystyle{ACM-Reference-Format}
\balance
\bibliography{references}

\appendix


\section{Data Release}
\label{app:data-release}
The datasets can be downloaded from the project website at \url{https://lead-lag-forecasting.github.io}.\\

\textbf{\arxiv.}
The original web logs contained fine grained access information by individual users; our time series data aggregates these patterns to avoid privacy concerns.
In addition, there are two related concerns:

\begin{enumerate}
  \item \textbf{Ethics:} the arXiv organization has not historically made information about paper views/downloads public. 
  There is a ``neutrality policy'' against appearing to promote specific preprints on the site.
  By removing preprint identifiers from the time series, we address this concern.
  \item \textbf{Re-identification:} The access time series are linked with citations, which are publicly available.
  In principle, this enables (partial) re-identification of the preprints.
  Agreement to responsible usage will reduces this risk while still allowing large-scale scholarly analysis.
\end{enumerate}

\textbf{\github.} We also release the GitHub activities data used in
our study. These files are publicly readable without credentialization.


\section{Additional Details about the \arxiv Dataset}
\label{app:arxiv-details}

\begin{figure*}[htbp]

    \centering
    \hfill
    \begin{minipage}[b]{0.25\textwidth}
        \includegraphics[width=\linewidth]{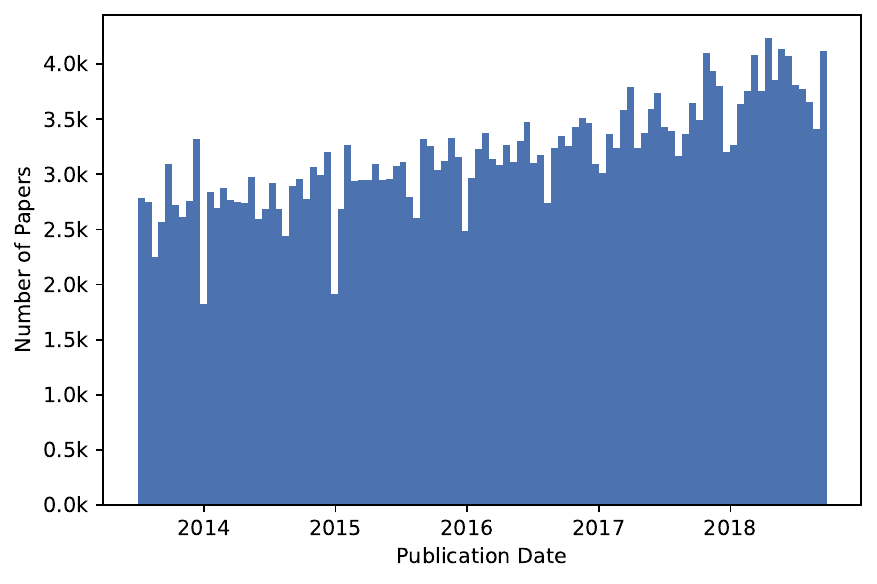}
    \end{minipage}
    \hfill
    \begin{minipage}[b]{0.25\textwidth}
        \includegraphics[width=\linewidth]{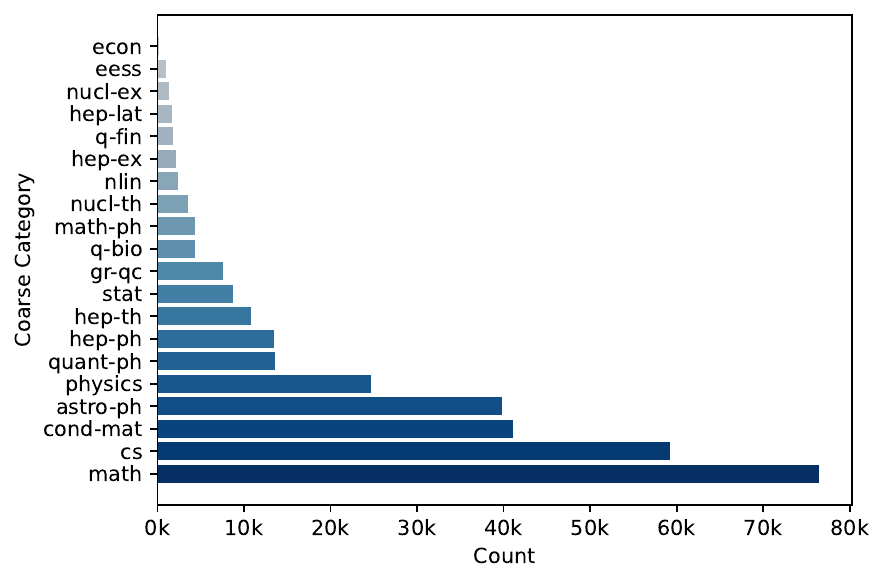}
    \end{minipage}
    \hfill
    \begin{minipage}[b]{0.25\textwidth}
        \includegraphics[width=\linewidth]{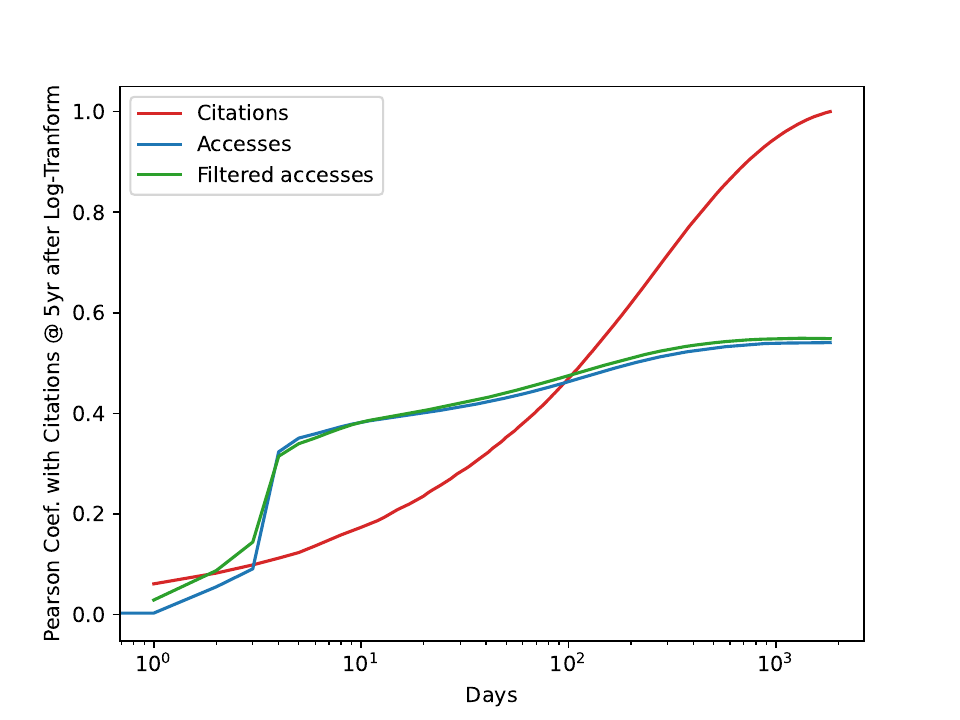}
    \end{minipage}
    \hfill
    \caption{\textbf{Left:} Histogram of paper publication time in the \texttt{train} split. \textbf{Middle:} Distribution across coarse categories in the \texttt{train} split. \textbf{Right:} Pearson correlation of citations (red), unfiltered accesses (blue), and filtered accesses (green) with 5-year citation count. Filtering is done by removing accesses from anonymized users with $>5000$ total accesses or $>50$ average daily accesses during their active periods.}
    \label{fig:coarse_category}
\end{figure*}

\textbf{Distribution.}
The distribution of paper publication dates is illustrated in Figure~\ref{fig:coarse_category} (Left). There are 152 fine-grained and 20 coarse first-choice categories.
\autoref{fig:coarse_category} (Middle) plots the coarse distribution for the train set.

\textbf{Bots Analysis.} The raw arXiv data has undergone a conservative  bot filtering process based on user access frequencies internally by \arxiv. To explore how bots affect the predictive power of accesses for citations, we conducted a second stage filtering approach by removing accesses from anonymized users with $>5000$ total accesses or $>50$ average daily accesses during their active periods, amounting to 0.2\% of users. The second-stage filtered accesses exhibit very similar patterns to first-stage filtered accesses when we measure their Pearson correlations with 5-year citations in Fig.~\ref{fig:coarse_category} (Left), suggesting that second processing hardly affects in the citation prediction task. For this reason, we proceeded to release the raw data we received from \arxiv, conservatively filtered by \arxiv.

\textbf{User Identification and Access Counting.} 
The user identifiers in the pseudonymized raw data are derived from the IP address of the request (represented as 9 hex digits from an MD5 hash), ignoring cookies or login status. Consequently, users with different IP addresses are assigned different IDs, while distinct users sharing an IP address are assigned the same ID. While this IP-based method introduces potential distortions and a small probability of hash collisions, it serves as a robust baseline for usage tracking. Crucially, our aggregation methodology limits usage counting to at most one access per user per day; multiple requests from the same ID within a 24-hour window are collapsed into a single access event.

\begin{table*}[htbp]
\centering
\scriptsize
\caption{\textbf{Percentages of papers with data anomalies in the \arxiv dataset.}}
\resizebox{0.9\textwidth}{!}
{%
\begin{tabular}{lcccc}
\toprule
\multirow{2}{*}{\textbf{Anomaly}} 
& \textbf{Submitted before daily precision} & \textbf{Daily precision labeled} & \textbf{Daily precision unlabeled} & \textbf{Entire dataset}\\
& $<$ \texttt{2013-07-01} (37\%) & \texttt{2013-07-01} - \texttt{2018-09-24} (25\%) & $>$ \texttt{2018-09-24} (38\%) & \\
\midrule
Missing \texttt{cumulative\_citations\_offset} data & 13.73\% & 11.15\% & 20.10\% & 15.51\% \\
\texttt{first\_access > 4}                          & 79.01\% & 2.97\% & 4.43\% & 31.45\% \\
\texttt{first\_citation < 0}                        & 16.89\% & 20.17\% & 17.52\% & 17.96\% \\
\texttt{first\_publication < 0}                     & 7.02\% & 7.87\% & 7.06\% & 7.25\% \\
\bottomrule
\end{tabular}}
\label{tab:arxiv-anomalies}

\end{table*}

\textbf{Dataset Features.} Below we provide a description of each key present in the \arxiv dataset:
\begin{itemize}
    \item \verb|first_publication|: difference in days between the publication date of the first version of the paper and arXiv submitted date;
    \item \verb|first_access|: difference in days between the date of the first access and arXiv submitted date;
    \item \verb|first_citation|: difference in days between the date of the first citation and arXiv submitted date;
    \item \verb|cumulative_accesses|: cumulative number of accesses from the arXiv submitted date, e.g., \verb|[1, 1, 2, ..., 3, 3]|;
    \item \verb|cumulative_citations_offset|: cumulative number of accesses from the arXiv submitted date, subtracting the initial citations on Day 0;
    \item \verb|initial_citations|: total number of citations before and on the arXiv submitted date.
\end{itemize}

\verb|initial_citations| can be nonzero due to a mismatch between the actual citation date and our inferred date based on the first publication of the citing papers: most of the recorded citations before the submitted date actually come from later versions of the citing papers. In order to avoid exposing too much later signal, we subtracted the initial citations from cumulative citation counts to form \verb|cumulative_citations_offset|. \verb|cumulative_citations_offset| can be empty for a paper due to missing citation data. Other anomalies include late first access (after day 4 \footnote{Most papers gain immediate access after being published on arXiv. However, some are not immediately released after submission. Therefore, we used a threshold of 4 to quantify late first access.}) and publication before submitted to arXiv (\autoref{tab:arxiv-anomalies}).


\textbf{Additional Details on Dataset Splits.} The analysis and benchmark model training are performed over the \verb|train| split of the \arxiv dataset. The benchmark evaluations are conducted on the \verb|test| split. We use the \verb|cumulative_accesses| key for the accesses sequences, and \verb|cumulative_citations_offset| for the citations sequences. 

The \verb|train_extra| split contains papers that are filtered out because of the following anomalies:
\begin{itemize}
    \item Papers that are not in the citation data obtained from the Semantic Scholar Datasets API (16\% filtered out). 
    \item Papers that are published before or on \verb|2013-06-30|, as daily access records are not available before this date (37\% from the previous step filtered out). Another 36\% of papers after the previous filtering have histories shorter than 5 years and are put into the \verb|unlabelled| test split.
    \item Papers with first access later than day 4 since the date of submission to arXiv,
    as this suggests that the preprint has abnormal access sequence with little early signal (2\% from the previous step filtered out).
\end{itemize}


\section{Additional Details about the \github Dataset}
\label{app:github-details}
\begin{figure}[htbp]
    \centering
        \includegraphics[width=.6\linewidth]{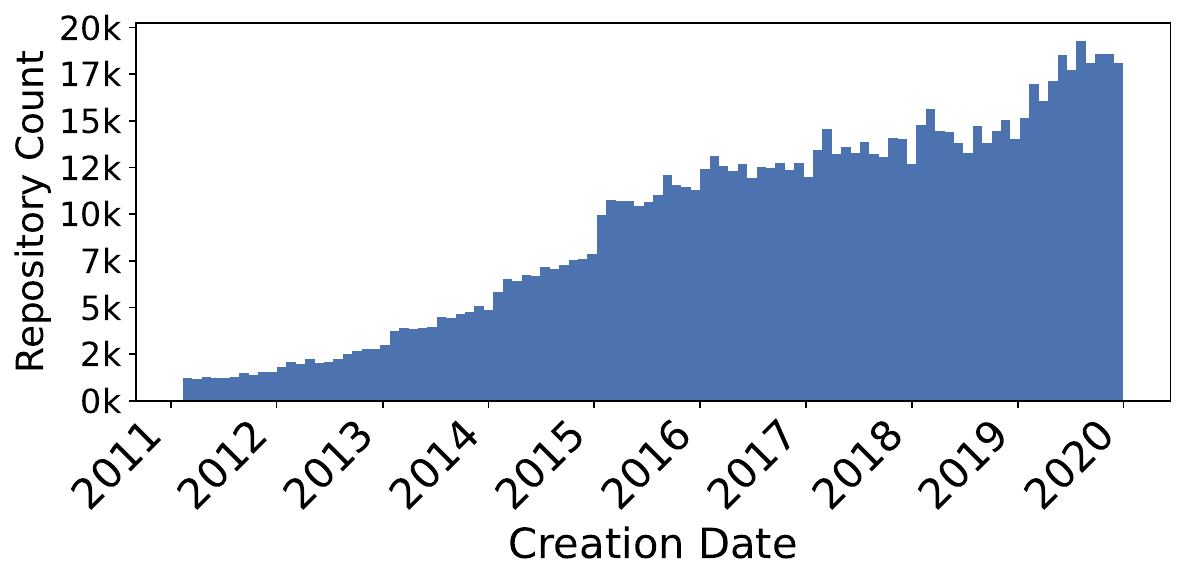}

        \includegraphics[width=.6\linewidth]{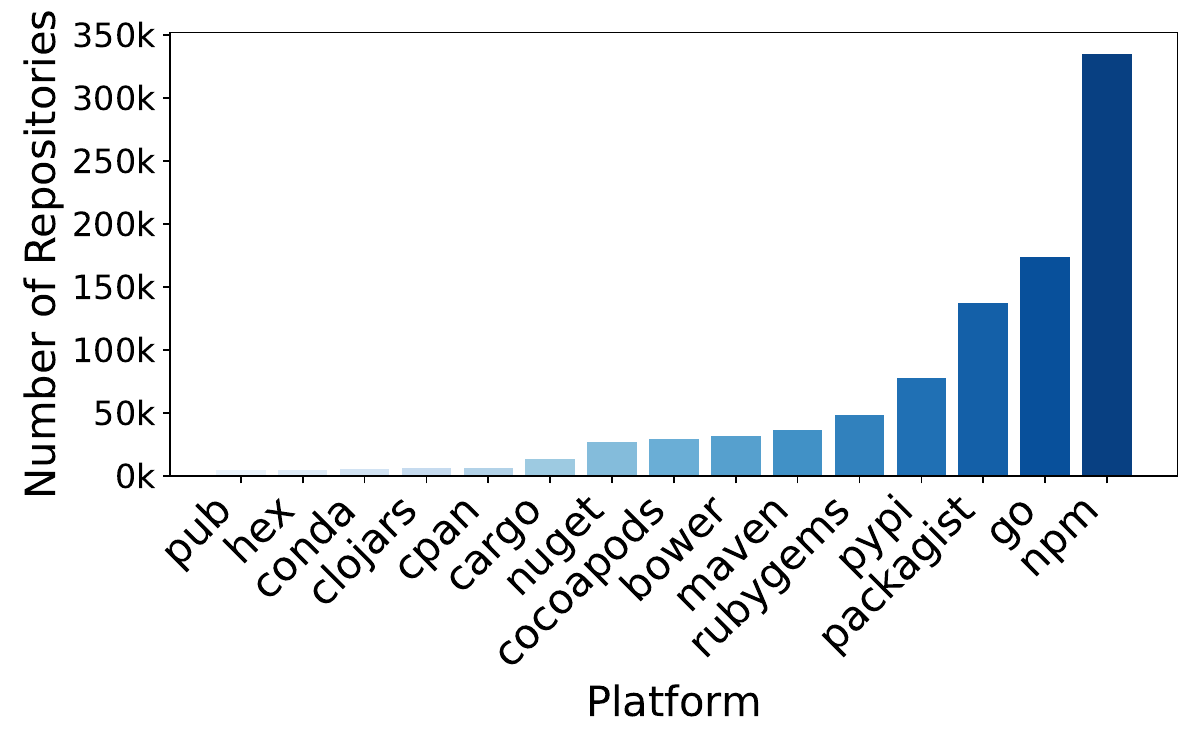}

    \caption{\textbf{Top:} GitHub Repository creation timeline. \textbf{Bottom:} Distribution across coarse packages.}
    \label{fig:plaform-dist}
\end{figure}

\textbf{Distribution.}
\autoref{fig:plaform-dist} shows the creation timeline of GitHub repositories (top) in the dataset and the distribution of top 15 packages (bottom).
\textbf{Dataset Features.} Below we provide a description of each key present in the \github dataset:
\begin{itemize}
    \item \verb|id|: unique identifier of the GitHub repository, as a string of \verb|{owner_name}/{repo_name}|;
    \item \verb|platform|: list of name strings of platforms on which the repository is hosted, e.g., \verb|[`pypi', `npm']|;
    \item \verb|created_date|: date of the repository creation recorded by Ecosyste.ms;
    \item \verb|first_create|: difference in days between the date of the first CreateEvent (i.e., repository, branch, or tag creation) and \verb|created_date|.
    \item \verb|first_push|: difference in days between the date of the first PushEvent and \verb|created_date|;
    \item \verb|first_star|: difference in days between the date of the first WatchEvent (star) and \verb|created_date|;
    \item \verb|first_fork|: difference in days between the date of the first ForkEvent and \verb|created_date|;
    \item \verb|cumulative_pushes_w_creates|: cumulative number of PushEvents (pushes) plus CreateEvents indexed by day from \verb|created_date|, e.g., \verb|[1, 1, 2, ..., 3, 3]|;
    \item \verb|cumulative_pushes|: cumulative number of PushEvents (pushes) indexed by day from \verb|created_date|;
    \item \verb|cumulative_stars|: cumulative number of WatchEvents (stars) indexed by day from \verb|created_date|;
    \item \verb|cumulative_forks|: cumulative number of ForkEvents (forks) indexed by day from \verb|created_date|.
\end{itemize}

In principle, \verb|first_create| should be 0 as the first CreateEvent coincides with repository creation. However, it is nonzero for 35\% of data, due to the repository's first CreateEvent missing in GH Archive or not matching the repository creation recorded by Ecosyste.ms. There can also be other data anomalies (\autoref{tab:github-anomalies}), such as events before \verb|created_date| due to repository renaming.

\begin{table}[!h]
\centering
\scriptsize
\caption{\textbf{Percentages of package repositories with data anomalies in the \github dataset.}}
\resizebox{.5\textwidth}{!}
{%
\begin{tabular}{lccc}
\toprule
\multirow{2}{*}{\textbf{Anomaly}} 
& \textbf{Labeled} & \textbf{Unlabeled} & \textbf{Entire dataset}\\
& < \texttt{2020-01-01} (48\%) & $\geq$ \texttt{2020-01-01} (52\%) & \\
\midrule
\texttt{first\_create != 0}  & 36.69\% & 34.21\% & 35.41\% \\
\texttt{first\_push < 0}     & 2.05\% & 2.30\% & 2.18\% \\
\texttt{first\_star < 0}     & 0.56\% & 0.53\% & 0.54\% \\
\texttt{first\_fork < 0}     & 0.36\% & 0.29\% & 0.32\% \\
\bottomrule
\end{tabular}}
\label{tab:github-anomalies}
\end{table}

\textbf{Additional Details on Dataset Splits.} The analysis and benchmark model training are performed over the \verb|train| split of the \github dataset. The benchmark evaluations are conducted on the \verb|test| split. While the cumulative count features are taken as prediction inputs and targets, the first event features can be used for filtering. We moved the examples (2\%) with negative \verb|first_push|, \verb|first_star|, or \verb|first_fork| to \verb|train_extra|.

\end{document}


\maketitle

\section{Additional Experimental Details}
\label{app:baselines}
\subsection{Classification}
In the following, for each horizon $t\!\in\!\{30,100,365\}$ for \arxiv and $t\!\in\!\{10,30,100\}$ for \github, and each threshold of $50$ for \arxiv and $10$ for \github we perform the following:

\textbf{One Feature Logistic Regression:} For each horizon $t$ we fit a logistic regressor on log-transformed cumulative counts \textit{at} time $t$:
\[
\begin{aligned}
\widehat{LR}^{(i)}_t &= \sigma\!\bigl(w^\top x^{(i)}_t+b\bigr), \\
x^{(i)}_t &=
\begin{cases}
\bigl[\text{log(cum.\ accesses)}^{(i)}_t\bigr] & (\arxiv)\\
\bigl[\text{log(cum.\ pushes)}^{(i)}_t,\ \text{log(cum.\ stars)}^{(i)}_t\bigr] & (\github)
\end{cases}
\end{aligned}
\]
The logistic regression from Scikit-Learn library, with balanced class weights and random seed $42$ is used to fit the model.

\textbf{Vanilla Logistic Regression:} For each horizon, we fit a logistic regressor on log-transformed cumulative counts \textit{up to} that horizon.

The regressor is implemented in Pytorch, using the loss function \texttt{BCEWithLogitsLoss}, with the positive weight set accordingly for each dataset ($11.24$ for \arxiv and $7.81$ for \github). Models are trained for $35$ epochs with a learning rate of $0.01$ and AdamW optimizer.

\textbf{Time-MoE logistic regression:}
For each entity, for a fixed input horizon $t$ we construct a feature vector by concatenating the daily cumulative counts in log space across the input horizon. We normalize the input with the global mean and standard deviation computed across all entities and time steps in the training set, separately for each channel. We pass each input through Time-MoE (large) \citep{shi2024time}, a mixture of experts foundation model with 200M active parameters to embed each input sequence.  
The regressor is implemented in Pytorch, using \texttt{BCEWithLogitsLoss}, with the positive weight set accordingly for each dataset ($11.24$ for \arxiv and $7.81$ for \github). Models are trained for $35$ epochs with a learning rate of $0.01$ and AdamW optimizer. For the results shown in the main body, we used the final hidden layer of the model along with mean pooling to extract the embeddings.


\textbf{InceptionTime:} We use the same input representation as vanilla logistic regression. The model consists of 6 inception layers. It was trained for 5 epochs with learning rate 1e-4, batch size 512, AdamW optimizer, cosine scheduler with 1-epoch warmup, and gradient clipping at 1. The last epoch checkpoint is used for evaluation on the test set. A single model is used without ensembling.

\subsection{Regression}
We provide a description of each baseline in the lag outcome regression task benchmarking below. Each model was trained on a single GPU. While our experiments were conducted using NVIDIA Titan RTX and NVIDIA GeForce RTX 3090 GPUs, the models are also compatible with GPUs that have smaller memory capacity.

\textbf{Mean of Early Years (MEY)}: MEY is a simple but strong baseline for citation prediction. For each paper, it calculates the average number of new citations observed at each time step within the input horizon, and uses this as the predicted number of new citations for each future time step. While some previous works~\citep{abrishami2019predicting} use yearly resolution, our implementation uses daily resolution as our longest input horizon is one year. MEY is a univariate method, so we use only the lag channel for prediction, excluding the lead channels. And as MEY only leverages each entity's own input horizon, predictions are directly made over the test set.

\textbf{K-Nearest Neighbors (KNN)}: For each entity, we construct a feature vector by concatenating the daily cumulative counts in log space across the input horizon for all lead and lag channels, i.e. accesses and citations for \arxiv, and pushes, stars and forks for \github. To make a prediction for a test entity, we identify its five nearest neighbors in the training set based on the L2 distance of the feature vectors, and use the median of the ground truth lag outcomes from these neighbors as the final prediction.

\textbf{Linear Regression}: Similar to KNN, the input variables consist of daily cumulative counts in log space across the input horizon for all lead and lag channels. The output variable is the lag outcome, i.e. cumulative count at year 5, in the log space. A linear regression model is fit on the training set, and then applied for prediction on the test set.

\textbf{Multi-Layer Perceptron (MLP)}: The input and output variables are similar to those of linear regression. Additionally, we normalize the input with the global mean and standard deviation computed across all entities and time steps in the training set, separately for each channel. We use a 3-layer MLP with hidden size 1024 and \verb|tanh| activations in-between. We perform hyperparameter search on the validation set. The final model is trained with the following hyperparameters: learning rate 1e-3, batch size 64, SGD optimizer, cosine scheduler with 1-epoch warmup, gradient clipping at 1, and 5 epochs. The last checkpoint is used for evaluation on the test set. We do not perform any early stopping, as validation set does not demonstrate noticeable overfitting behavior. 

\textbf{Transformer}: Each entity is represented as a sequence over the input horizon, where each time step is a vector containing the log-transformed cumulative counts for all channels. We apply the same input normalization as used for MLP. The architecture consists of learned positional encoding, followed by 3 attention blocks with bidirectional multi-head attention and pre-Layer Normalization. We use 4 attention heads, hidden size 256 and feed forward size 1024. To predict the lag outcome in the log space, we apply adaptive pooling across the sequence outputs, followed by a linear layer. Like the MLP model, we perform hyperparameter search over the validation set. The final model is trained with the following hyperparameters: learning rate 1e-4, batch size 512, AdamW optimizer, cosine scheduler with 1-epoch warmup, gradient clipping at 1, and 5 epochs. The last checkpoint is used for evaluation on the test set; we do not perform any early stopping.

\textbf{Time-MoE K-Nearest Neighbors (Time-MoE KNN)}: Each entity is same as transformer with the same normalizations. The entities are fed to the Time-MoE~\citep{shi2024time} and embeddings of sequences up to each input horizon are used as features for the KNN. We show ablations for embeddings from the middle decoder layers and last hidden layer. We also show ablations of taking the last token, and average and max pooling along the token axis. To make a prediction for a test entity, we identify its five nearest neighbors in the training set based on the L2 distance of the feature vectors, and use the median of the ground truth lag outcomes from these neighbors as the final prediction. The results using mean pooling + final layer are shown in the body of the paper for both \arxiv and \github.

\begin{figure*}[htbp]
    \includegraphics[width=0.9\linewidth]{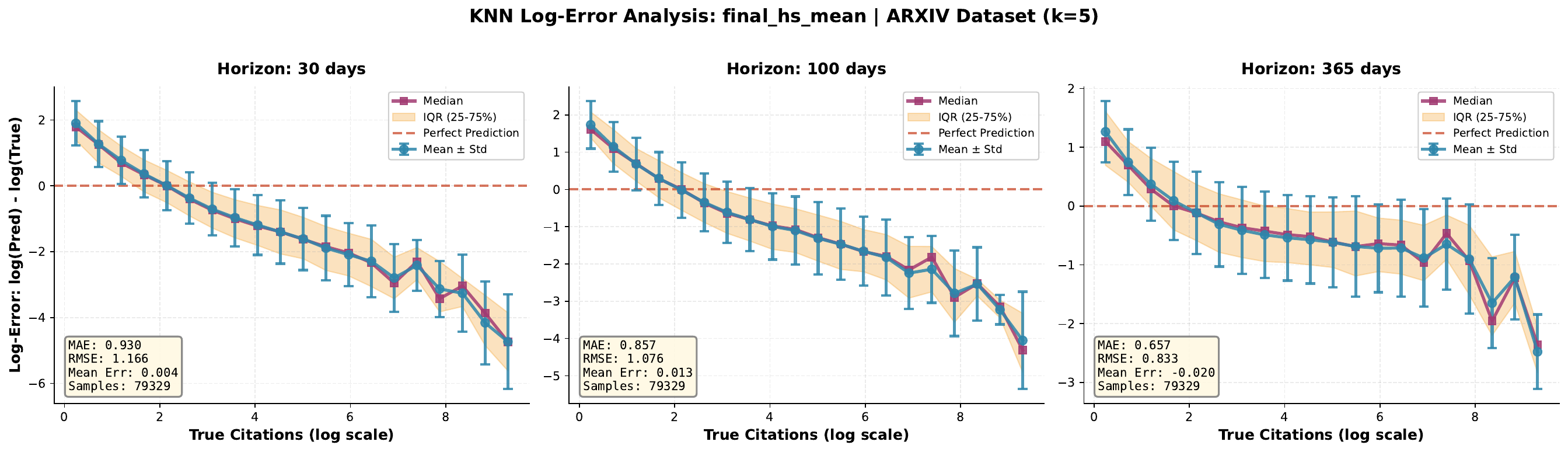}
    \caption{Binned error analysis showing how prediction accuracy varies with true citation count (in log-scale) for Time-MoE KNN citation forecasting. Each panel displays log-error trends across three horizons using Time-MoE features (k=5). Blue error bars indicate mean one standard deviation. The dashed red line represents perfect prediction (zero error). Across all horizons, the model exhibits systematic underestimation for low-citation papers (positive log-error at low true citation counts) and overestimation for highly-cited papers (negative log-error at high counts), with median errors converging toward zero as citation counts increase. This bias pattern is most pronounced at the 30-day horizon and diminishes at longer horizons. The 365-day horizon shows the best performance with MAE=0.657 and narrower error distributions, particularly for papers with moderate citation counts (log-scale 2-6). Prediction uncertainty (error spread) higher for very high citation papers across all horizons.}
    \label{fig:pop-arxiv}
\end{figure*}

\begin{figure*}[htbp]
    \includegraphics[width=0.9\linewidth]{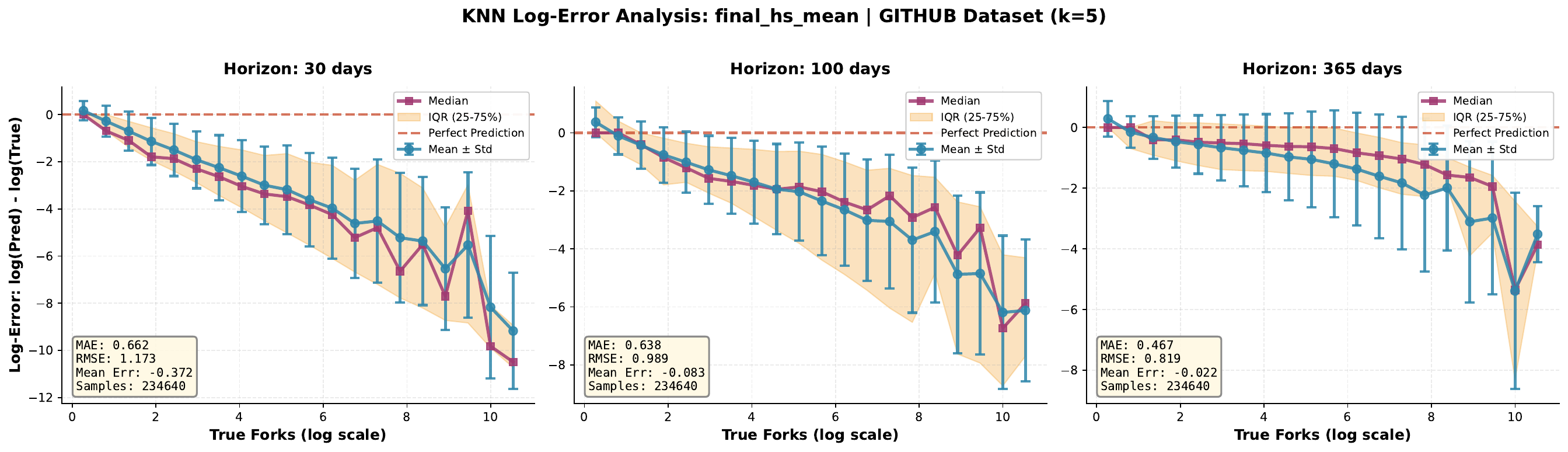}
    \caption{Binned error analysis showing how prediction accuracy varies with true number of forks count (in log-scale) for Time-MoE KNN forks forecasting. Across all horizons, the model exhibits systematic underestimation for low-forked repositories (positive log-error at low true forks counts) and overestimation for highly-forked repositories (negative log-error at high counts), with median errors converging toward zero as forks counts increase. This bias pattern is most pronounced at the 30-day horizon and diminishes at longer horizons. The 365-day horizon shows the best performance with MAE=0.467 and narrower error distributions, particularly for repositories with moderate fork counts (log-scale 2-8). Prediction uncertainty (error spread) is higher for very high forked repositories across all horizons. The errors observed here are larger than their counterparts for \arxiv in \autoref{fig:pop-arxiv}, due to very sparse nature of \github dataset.}
    \label{fig:pop-github}
\end{figure*}

\section{Discussion of Forecasting Limitations}
\label{app:forecasting-limitations}
\begin{figure*}[htbp]
    \includegraphics[width=\linewidth]{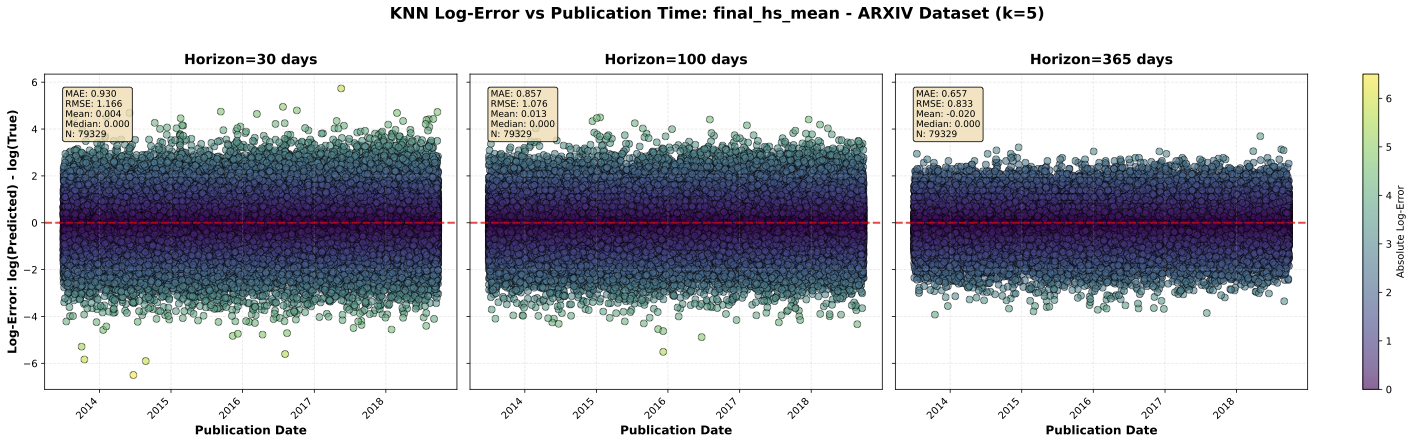}
    \caption{Temporal distribution of prediction errors for Time-MoE KNN across three horizons on the ArXiv dataset Each panel shows log-error (log(Predicted) - log(True)) as a function of release date, with point color intensity indicating absolute error magnitude. The horizontal red line at zero represents perfect prediction. No systematic temporal trends are apparent, suggesting stable prediction performance across the publication period. Performance metrics improve substantially with longer horizons.}
    \label{fig:pub_arxiv}
\end{figure*}
\begin{figure*}[htbp]
    \includegraphics[width=\linewidth]{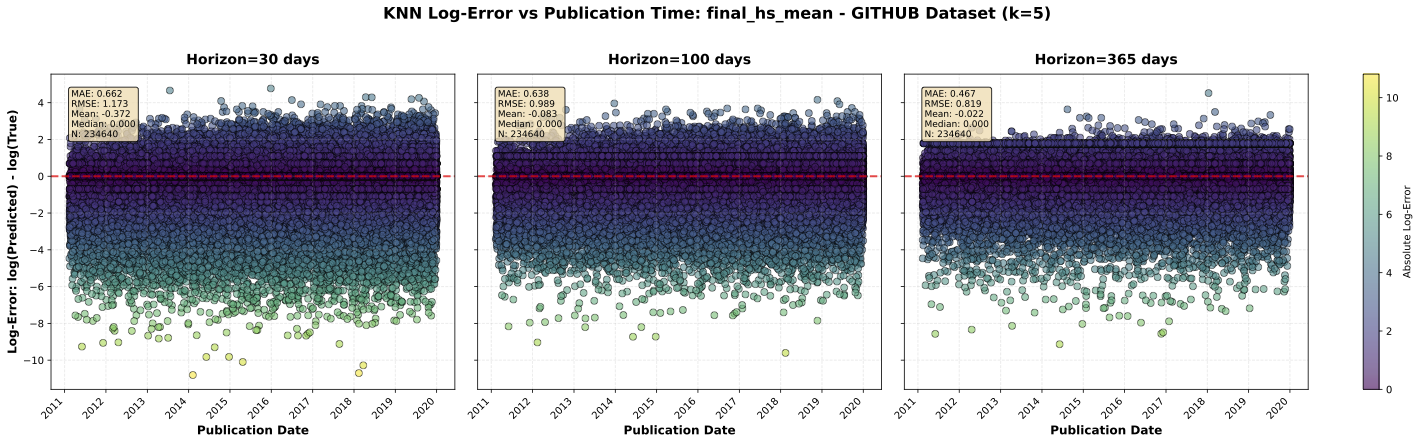}
    \caption{Temporal distribution of prediction errors for Time-MoE KNN across three horizons on the \github dataset Each panel shows log-error (log(Predicted) - log(True)) as a function of release date, with point color intensity indicating absolute error magnitude. The horizontal red line at zero represents perfect prediction. No systematic temporal trends are apparent, suggesting stable prediction performance across the publication period. Performance metrics improve substantially with longer horizons.}
    \label{fig:pub_github}
\end{figure*}

Our datasets introduce exciting challenges beyond standard forecasting in several aspects:

\textbf{Cross-series Generalization}: With 2.3M papers and 3M repos, models need to be able to predict for unseen entities, requiring broad pattern recognition beyond entity-specific trends.

\textbf{Cross-channel Prediction}: Models must map early signals (accesses or stars/pushes) to different outcome types (citations or forks), not just extrapolate the same variable. 

\textbf{Temporal Gap and Sparsity:} A long gap (30–365 day input $\rightarrow$ 5-year target) means the lag signal is often sparse or zero early on, forcing models to find predictive signals without short-term cues.

\textbf{Long-tailed distributions}: In our datasets, citations and forks follow power-law distributions. Only a small percentage achieve high impact, so models must learn from rare events without overfitting to common low-impact cases.

Current forecasting methods face two primary limitations when applied to lead-lag forecasting (LLF):

\paragraph{Cross-Series Generalization Constraints.} 
Prevailing models are typically designed for forecasting future values within the same set of series, while LLF requires prediction for entirely unseen series.
\begin{itemize}
    \item \textbf{Training/Testing Paradigm Mismatch:} Existing approaches assume chronological splits across all series (e.g., training on months 1--10, testing on months 11--12 for every energy station in the ETT dataset~\citep{zhou2021informer}). In LLF, training and testing instead involve disjoint series (e.g., training on papers published up to 5 years ago, testing on entirely new papers), which current architectures are not designed to handle.
    \item \textbf{Scalability Constraints:} Most of the sophisticated models such as the Informer~\citep{zhou2021informer} treat all variates within a time window as a single sample. With 1M+ variates in our datasets, this approach becomes computationally infeasible and architecturally rigid, since it prevents accommodating new variates at inference.
\end{itemize}

\paragraph{Temporal Prediction Scope Mismatches.} 
\begin{itemize}
    \item \textbf{Long-Horizon Accumulation Error:} Conventional point-process models such as the Transformer Hawkes~\citep{zuo2020transformer} forecast only short horizons (1--96 steps ahead). Extending to 5-year predictions ($\approx1825$ steps) requires iterative forecasting, where compounding errors undermine reliability.
    \item \textbf{Periodicity Assumptions:} Architectures such as Autoformer~\citep{wu2021autoformer} rely on capturing regular seasonal patterns. However, long-term dynamics in LLF do not exhibit such periodicity, rendering these assumptions ineffective.
\end{itemize}

\section{Ablations}

In this section, we perform further ablations to ensure model and dataset robustness:



\textbf{Feature Robustness Ablations.} Below, we perform further ablations for Time-MoE KNN model, using final hidden layer and mean pooling, same setting as main paper results, to check for robustness of our models to different features. \autoref{tab:timemoe_ablation} shows the results for \arxiv and \autoref{tab:timemoe_ablation_github} for \github. We observe that our model is fairly robust to dataset features across both datasets. 

\textbf{Popularity Robustness.}
Binned error analysis showing how prediction accuracy varies with true citation count (in log-scale) for Time-MoE KNN citation forecasting. Each panel displays log-error trends across three horizons using Time-MoE features (k=5). Across all horizons, the model exhibits systematic underestimation for low-citation papers (positive log-error at low true citation counts) and overestimation for highly-cited papers (negative log-error at high counts), with median errors converging toward zero as citation counts increase. This bias pattern is most pronounced at the 30-day horizon and diminishes at longer horizons.  For \arxiv in \autoref{fig:pop-arxiv}, the 365-day horizon shows the best performance with MAE=0.657 and narrower error distributions, particularly for papers with moderate citation counts (log-scale 2-6). Prediction uncertainty (error spread) increases for both very low and very high citation papers across all horizons. \github in \autoref{fig:pop-github} shows similar results overall, with larger errors due to very sparse nature of the dataset.

\textbf{Release/Publication Date Robustness.}
Below we show temporal distribution of prediction errors for k-nearest neighbors (k=5) citation forecasting. using the final hidden layer and mean pooling embeddings of Time-MoE as similarity metric across three horizons on the ArXiv dataset (N=79,329 papers, 2014-2018) in \autoref{fig:pub_arxiv} and \github dataset (N=234640, 2011-2020) in \autoref{fig:pub_github}. No systematic temporal trends are apparent, suggesting stable prediction performance across the publication period. Performance metrics improve substantially with longer horizons: MAE decreases from 0.930 (30 days) to 0.857 (100 days) to 0.657 (365 days), while RMSE follows a similar trend. The median error remains near zero across all horizons, indicating unbiased predictions on average. Error variance decreases notably at longer horizons, as evidenced by the tighter clustering of points around the zero line in the 365-day panel. 

\begin{figure*}[htbp]
    \centering
    \includegraphics[width=.55\linewidth]{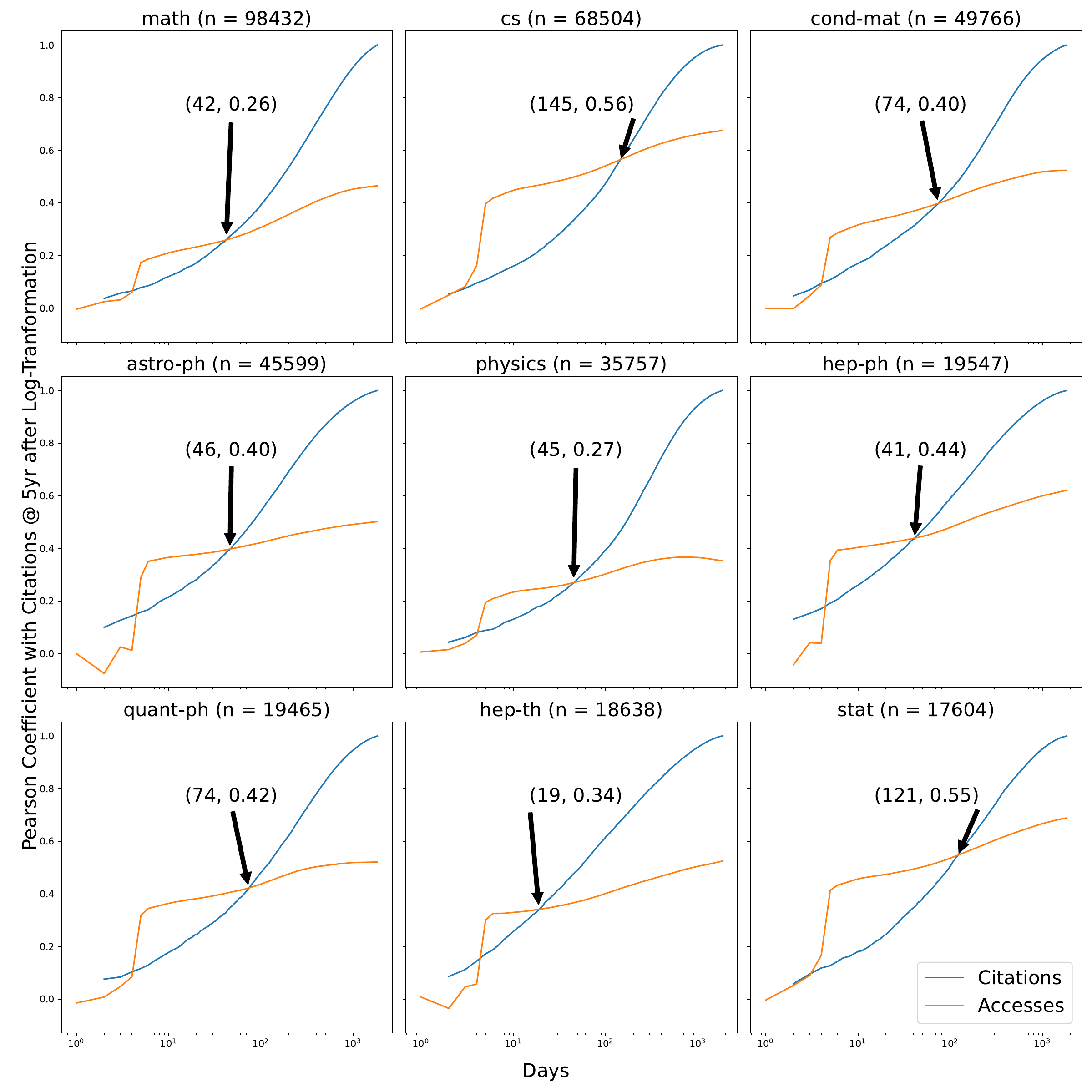}
    \includegraphics[width=.8\linewidth]{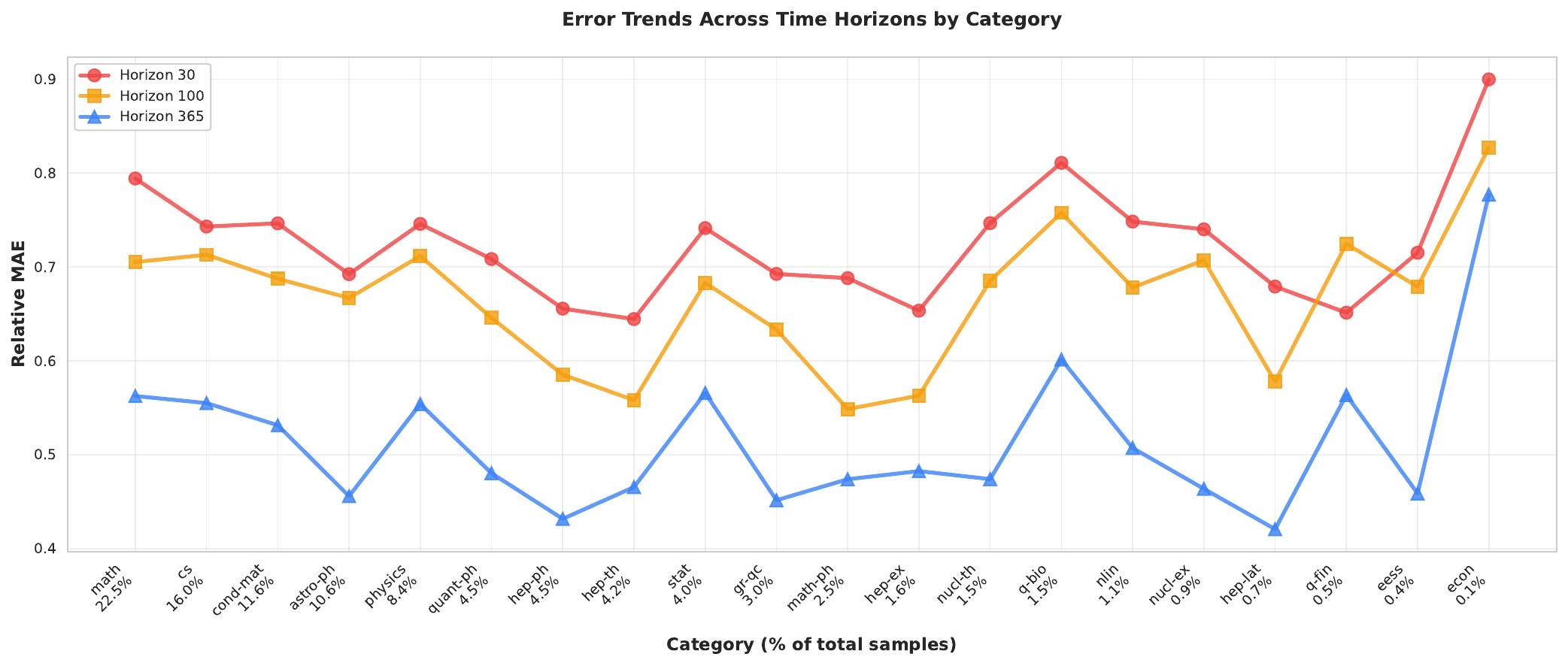}
    \caption{Analysis of citation prediction across different categories on the \arxiv dataset. \textbf{Top}: Pearson correlation of citations (red) and accesses (blue) with 5-year citation count for papers in each top arXiv categories. Access data shows strong early correlation for all categories. \textbf{Bottom}: Relative Mean Absolute Error (MAE) across scientific categories, showing considerable variation in prediction difficulty. Categories are ordered by dataset representation, with mathematics (22.5\%) and computer science (16.0\%) being most prevalent. While very underrepresented categories like econ (0.1\%) tend to perform worse, error patterns do not reveal large std of errors across different categories. Shorter horizons (30 days) consistently exhibit higher relative errors across all categories, while longer horizons (365 days) achieve substantially better performance. Error patterns reveal that shorter horizons (30 days) consistently exhibit higher relative errors across all categories, while longer horizons (365 days) achieve substantially better performance.}
    \label{fig:cat-arxiv}
\end{figure*}
\textbf{Category Ablations.}
Below we show analysis of citation prediction across different scientific categories, on the ArXiv dataset. \autoref{fig:cat-arxiv}(top) shows Pearson correlation of citations (red) and accesses (blue) with 5-year citation count for papers in each top arXiv categories. Access data shows strong early correlation for all categories.
\autoref{fig:cat-arxiv}(bottom) shows performance analysis of Time-MoE KNN for three time horizons (30, 100, and 365 days), showing moderate variation in prediction difficulty. Categories are ordered by dataset representation, with mathematics (22.5\%) and computer science (16.0\%) being most prevalent. While very underrepresented categories like econ (0.1\%) tend to perform worse, error patterns generally do not reveal systematic performance gaps across different categories.  Shorter horizons (30 days) consistently exhibit higher relative errors across all categories, while longer horizons (365 days) achieve substantially better performance.

\begin{table*}[htbp]
\centering
\scriptsize
\caption{\textbf{Ablation study results for Time-MoE KNN on the \arxiv test split.}}
\resizebox{\textwidth}{!}
{%
\begin{tabular}{llccc ccc ccc}
\toprule
\multirow[c]{2}{*}{\textbf{Pooling Method}} & \multirow[c]{2}{*}{\textbf{Layer}} & \multicolumn{3}{c}{\textbf{30 Days}} & \multicolumn{3}{c}{\textbf{100 Days}} & \multicolumn{3}{c}{\textbf{365 Days}} \\
\cmidrule(lr){3-5} \cmidrule(lr){6-8} \cmidrule(lr){9-11}
 &  & MAE-log & MAE & MAPE & MAE-log & MAE & MAPE & MAE-log & MAE & MAPE \\
\midrule
\multirow{2}{*}{\textbf{Max Pooling}} & Middle Layer & 0.933 & 19.005 & 0.742 & 0.861 & 17.686 & 0.686 & 0.649 & 12.822 & 0.504 \\
 & Last Layer & 0.930 & 19.018 & \textbf{0.731} & 0.867 & 17.982 & 0.693 & 0.660 & 13.307 & 0.533 \\
\multirow{2}{*}{\textbf{Mean Pooling}} & Middle Layer & 0.928 & 19.011 & 0.734 & 0.852 & 17.577 & 0.682 & 0.649 & 12.977 & 0.506 \\
 & Last Layer & 0.931 & 18.954 & 0.732 & 0.857 & 17.613 & 0.686 & 0.657 & 13.232 & 0.521 \\
\multirow{2}{*}{\textbf{Last Token}} & Middle Layer & \textbf{0.926} & 18.877 & 0.738 & \textbf{0.851} & 17.349 & 0.677 & \textbf{0.647} & \textbf{12.758} & \textbf{0.498} \\
     & Last Layer & 0.927 & \textbf{18.856} & 0.740 & 0.847 & \textbf{17.158} & \textbf{0.670} & 0.657 & 13.195 & 0.508 \\
\bottomrule
\end{tabular}}
\label{tab:timemoe_ablation}
\end{table*}

\begin{table*}[htbp]
\centering
\scriptsize
\caption{\textbf{Ablation study results for Time-MoE KNN on the \github test split.}}
\resizebox{\textwidth}{!}
{%
\begin{tabular}{llccc ccc ccc}
\toprule
\multirow[c]{2}{*}{\textbf{Pooling Method}} & \multirow[c]{2}{*}{\textbf{Layer}} & \multicolumn{3}{c}{\textbf{30 Days}} & \multicolumn{3}{c}{\textbf{100 Days}} & \multicolumn{3}{c}{\textbf{365 Days}} \\
\cmidrule(lr){3-5} \cmidrule(lr){6-8} \cmidrule(lr){9-11}
 &  & MAE-log & MAE & MAPE & MAE-log & MAE & MAPE & MAE-log & MAE & MAPE \\
\midrule
\multirow{2}{*}{\textbf{Max Pooling}} & Middle Layer & \textbf{0.666} & \textbf{11.88} & 0.926 & \textbf{0.552} & \textbf{10.90} & 0.877 & \textbf{0.347} & \textbf{8.38} & 0.658 \\
 & Last Layer & 0.675 & 12.00 & 0.929 & 0.558 & 11.08 & 0.886 & 0.356 & 8.75 & \textbf{0.656} \\
\multirow{2}{*}{\textbf{Mean Pooling}} & Middle Layer & 0.696 & 11.99 & 0.930 & 0.633 & 11.11 & \textbf{0.855} & 0.351 & 8.75 & 0.678 \\
 & Last Layer & 0.716 & 12.05 & 0.925 & 0.578 & 11.11 & 0.872 & 0.355 & 8.92 & 0.686 \\
\multirow{2}{*}{\textbf{Last Token}} & Middle Layer & \textbf{0.666} & 11.93 & \textbf{0.918} & 0.561 & 11.13 & 0.867 & 0.404 & 9.36 & 0.707 \\
 & Last Layer & 0.712 & 12.10 & \textbf{0.918} & 0.565 & 11.25 & 0.877 & 0.378 & 9.58 & 0.703 \\
\bottomrule
\end{tabular}}
\label{tab:timemoe_ablation_github}
\end{table*}

\bibliographystyle{ACM-Reference-Format}
\bibliography{references}